\documentclass{article}

\usepackage[preprint]{neurips_2026}
\usepackage{float}
\usepackage[utf8]{inputenc}
\usepackage[T1]{fontenc}
\usepackage{hyperref}
\usepackage{url}
\usepackage{booktabs}
\usepackage{amsfonts}
\usepackage{amsmath}
\usepackage{amssymb}
\usepackage{amsthm}
\usepackage{nicefrac}
\usepackage{microtype}
\usepackage{xcolor}
\usepackage{graphicx}
\usepackage{algorithm}
\usepackage{algpseudocode}
\usepackage{multirow}
\usepackage{caption}
\usepackage{subcaption}
\usepackage{mathtools}
\usepackage{wrapfig}
\usepackage{tikz}
\usetikzlibrary{arrows.meta, positioning, calc}

\newcommand{\piZ}{\pi_0}
\newcommand{\piOne}{\pi_1}

\newcommand{\E}{\mathbb{E}}
\newcommand{\R}{\mathbb{R}}
\newcommand{\norm}[1]{\left\|#1\right\|}

\title{%
  Divergence-Suppressing Couplings for Rectified Flow
}

\author{%
  Yimeng Min \qquad Carla P. Gomes \\
  Department of Computer Science\\
  Cornell University\\
  Ithaca, NY 14850 \\
  \texttt{\{min,gomes\}@cs.cornell.edu} \\
}

\begin{document}

\maketitle

\begin{abstract}
The promise of Rectified Flow rests on producing self-generated
couplings whose trajectories are straight, or nearly so. In practice,
trajectories generated by the base flow model can bend and intertwine,
and the resulting coupling inherits this distortion. In this paper,
we identify that such trajectory entanglement is often associated
with regions of nonzero divergence in the learned velocity field,
where local expansion or contraction distorts trajectories and steers
particles away from their ideal endpoints. We then propose
divergence-suppressing couplings for Rectified Flow, an
offline correction that attenuate the divergent component of the
learned velocity during coupling generation. The
correction is paid only once per coupling pair and amortized over
training, so deployment runs plain Euler at identical wall-clock cost
to standard Rectified Flow. Empirically, this offline modification
yields consistent improvements on 2D synthetic benchmarks
and on image generation.
\end{abstract}

\section{Introduction}
\label{sec:intro}

Generative models based on continuous normalizing flows
\citep{chen2018neuralode,grathwohl2018ffjord} learn to transport a
simple source distribution $\pi_0$ (typically a Gaussian) to a complex
target distribution $\pi_1$ by integrating an ordinary differential
equation $\dot{x} = v_\theta(t, x)$ defined by a learned, time-dependent
velocity field $v_\theta$. Until recently, training such flows
required differentiating through the ODE solver via the
adjoint method, an expensive procedure that limited their scalability
\citep{chen2018neuralode,grathwohl2018ffjord}.

Flow matching (FM) resolves this by replacing solver-in-the-loop
training with a simple regression objective~\citep{lipman2022flow,liu2022rectifiedflow,
albergo2022building}. Given a prescribed
probability path between $\pi_0$ and $\pi_1$, the most common choice is using the linear
interpolant $x_t = (1-t)x_0 + tx_1$ between independently drawn
$x_0 \sim \pi_0$ and $x_1 \sim \pi_1$. Here, the velocity field is trained
to predict the conditional vector field that generates that path,
typically the constant direction $x_1 - x_0$. The resulting Conditional
Flow Matching (CFM) loss is simulation-free, scales like standard
supervised learning, and unifies several earlier formulations including
denoising diffusion \citep{sohl2015deep,ho2020denoising,song2020score}
and stochastic interpolants \citep{albergo2025stochastic}. Flow
matching has since produced strong results across image synthesis
\citep{esser2024scaling}, audio generation
\citep{le2023voicebox}, video generation
\citep{polyak2024movie}, and molecular and protein design
\citep{bose2023se}.

A shared bottleneck across this family is inference cost. Although
training is simulation-free, sampling still requires numerical
integration of an ODE, typically tens to hundreds of neural function
evaluations (NFE) to reach competitive quality. The reason is geometric:
when $x_0$ and $x_1$ are paired independently, the conditional vector
field at any intermediate $x_t$ averages the directions $x_1 - x_0$ of
all source--target pairs that pass through $x_t$, and these directions
disagree wherever coupling distortion\footnote{Integral curves of a deterministic ODE cannot literally cross. We use ``coupling distortion'' loosely for two related phenomena: \emph{coupling crossings}, where pairs $(x_0, x_1)$ from $\mathcal{C}$ have intersecting linear interpolants with different directions, inflating the variance of $\mathbb{E}[x_1 - x_0 \mid x_t]$; and compression of neighbouring Euler trajectories under $v_\theta$.}. The learned marginal field is
therefore curved, and accurate integration demands many small Euler
steps. Reducing NFE without sacrificing quality is a central research
challenge, addressed by higher-order solvers
\citep{lu2022dpm,zhang2022fast}, distillation
\citep{salimans2022progressive,song2023consistency}, and
the trajectory-straightening approach we focus on here.

Rectified Flow (RectFlow) 
pursues the last of these by attacking the curvature at its source:
the independent coupling~\citep{liu2022rectifiedflow,liu2023instaflow}. Starting from a base flow model $v_\text{base}$
trained with the standard CFM loss, one generates a new \emph{coupling
dataset} $\mathcal{C}$ by running the model forward from
$x_0 \sim \pi_0$ to produce $x_1 \approx \mathrm{Euler}(v_\text{base}, x_0)$,
and then retrains on these self-generated pairs. Because the new pairs
are approximately aligned along ODE trajectories, they cross less
than the independent coupling did, and the next-round velocity field
is closer to linear---enabling few-step or single-step generation. The
same coupling-quality intuition has independently motivated several
related approaches, including optimal-transport CFM
\citep{tong2023improving,pooladian2023multisample}, consistency
training \citep{song2023consistency}. Despite its simplicity, RectFlow has a fundamental weakness: the coupling quality is limited by the quality of the base model's forward integration. The retrained model inherits these errors, and additional rectification rounds provide diminishing returns.

The goal of this paper is to introduce an \textit{offline} method to improve the coupling quality of RectFlow, which we identify as the fundamental bottleneck limiting its effectiveness. In particular, we show that coupling distortion is often associated with regions of nonzero divergence in the learned velocity field, where local expansion or contraction distorts the generated source–target pairs and limits improvement across rectification rounds. To address this, we introduce \textit{DS-RectFlow}, which replaces vanilla Euler with a divergence-suppressing Euler integrator during offline coupling generation. By attenuating the irrotational component of the velocity at each integration step, this produces cleaner couplings with reduced distortion and improved mode alignment, while leaving the training objective, the model architecture, and the inference procedure entirely unchanged. The cost of the projection is therefore paid only once during offline data generation, and inference remains plain Euler at identical wall-clock cost to standard RectFlow. Empirically, this simple offline modification restores the compounding effect that rectification promises but rarely delivers in practice, yielding substantial improvements in sample quality at low NFE and enabling near one-step generation across 2D synthetic benchmarks and image generation.

\section{Background}
\label{sec:background}

\subsection{Flow Matching}
\label{sec:fm}

Flow matching (FM) learns a time-dependent velocity field $v_\theta : [0,1] \times \R^d \to \R^d$ such that integrating the ODE $\dot{x} = v_\theta(t, x)$ from a simple source $\piZ$ (typically $\mathcal{N}(0, I)$) transports samples to the target distribution $\piOne$~\citep{lipman2022flow}. The Training uses the CFM objective.
Given pairs $(x_0, x_1)$ with $x_0 \sim \piZ$ and $x_1 \sim \piOne$, define the linear interpolant $x_t = (1-t) x_0 + t\, x_1$.
The target velocity is the direction $x_1 - x_0$, and the loss is:
\begin{equation}
  \mathcal{L}_\text{CFM}(\theta)
  = \E_{t \sim U[0,1],\, (x_0, x_1) \sim \mathcal{C}}\bigl[
      \norm{v_\theta(t, x_t) - (x_1 - x_0)}^2
    \bigr],
  \label{eq:cfm}
\end{equation}
where $\mathcal{C}$ is the coupling (joint distribution over pairs).
With independent coupling $\mathcal{C}_\text{ind}$ ($x_0$ and $x_1$ sampled independently), the marginals $\{x_t\}_{t\in[0,1]}$ follow prescribed paths, but individual trajectories distort heavily near $t \approx 0.5$ when the source and target have separated modes.

At inference, the ODE is integrated numerically with Euler's method for $N$ NFE:
\begin{equation}
  x_{t + \Delta t} = x_t + \Delta t \cdot v_\theta(t, x_t),
  \quad \Delta t = 1/N.
  \label{eq:euler}
\end{equation}

\subsection{Rectified Flow}
\label{sec:rectflow}

\citet{liu2022rectifiedflow} introduced RectFlow as an iterative
procedure for straightening the learned ODE. The motivating
observation is simple: if the source-to-target coupling is
\emph{undistorted}, the optimal velocity field is linear and a
single Euler step gives exact integration.

The algorithm proceeds in rounds. Let $v^{(0)} = v_\text{base}$
be a model trained with independent coupling. In round $k$, we sample
$x_0^{(i)} \sim \piZ$ and integrate under the previous round's model,
\begin{equation}
  x_1^{(i)} = \text{Euler}(v^{(k-1)},\, x_0^{(i)},\, N_\text{gen}),
  \label{eq:van_coupling}
\end{equation}
forming a new coupling dataset
$\mathcal{C}^{(k)}_\text{van} = \{(x_0^{(i)}, x_1^{(i)})\}$ on which
we retrain $v^{(k)}_\text{rect}$ with the CFM loss. Since each
$x_1^{(i)}$ lies (approximately) on an integral curve of $v^{(k-1)}$,
the pairs in $\mathcal{C}^{(k)}_\text{van}$ are approximately aligned
along ODE trajectories and contain fewer distortion than the
independent coupling. Each round therefore straightens the field
further, and after a few rounds $v^{(k)}_\text{rect}$ can generate
good samples at NFE as low as one.
The CFM loss (Eq.~\ref{eq:cfm}) is minimised by the conditional
expectation:
\begin{equation}
  v_\theta^*(t, x_t) = \E\bigl[x_1 - x_0 \mid x_t = x\bigr].
  \label{eq:cfm_opt}
\end{equation}
Under an undistorted coupling, every pair passing through a given
$x_t$ shares roughly the same direction $x_1 - x_0$. The conditional
expectation then has low variance, the learned field is nearly
straight, and Euler integration is accurate at a single step. Under
a distorted coupling, pairs with conflicting directions pass through
the same $x_t$ and the conditional expectation averages them,
producing a curved field that only many NFE can integrate
accurately. Rectified Flow exploits this by building
$\mathcal{C}^{(k)}_\text{van}$ from ODE trajectories of the previous
round's model, which in principle cannot cross because they are
solutions of a single ODE.
In practice, $\mathcal{C}^{(k)}_\text{van}$ is only \emph{approximately}
undistorted. The previous round's model $v^{(k-1)}$ was itself
trained on a coupling that was not perfectly undistorted, so the
field it defines still has regions of nonzero divergence where
trajectories spread apart or pile up. Euler integration under such a
field overshoots these regions and produces curved, distorted
trajectories instead of the clean transports that would land at the
ideal endpoints. The resulting $(x_0, x_1)$ pairs carry the
distortion forward as training signal, the next round inherits it,
and reflow faithfully straightens a transport that was already bent.
The gains from additional rectification therefore saturate quickly.

The bottleneck is not the reflow loss or the model capacity. It is
the quality of the forward integrator that generates the couplings.
What we need is an integrator that produces cleaner $(x_0, x_1)$
pairs from the same learned field by steering trajectories away from
the divergent regions responsible for the distortion. 
\section{Related Work}
Two recent works identify the divergence of the learned velocity field as a quantity worth controlling, but pursue this goal through different routes than ours. \cite{cha2026training} introduce the Flow Divergence Sampler (FDS), a training-free refinement of flow matching that proves the optimal CFM residual admits a closed-form expression in terms of $\nabla \cdot u_\theta(x,t)$, identifying divergence as a data-free surrogate for local trajectory ambiguity; FDS exploits it at sampling time by drawing Gaussian perturbations around the current state and selecting the candidate with lowest divergence before each solver step. \cite{huang2026improving} instead derive a PDE for the error $\epsilon_t = p_t - \hat{p}_t$ between exact and learned probability paths, and show via Duhamel's principle that the total variation gap is bounded by a loss combining $u_t - v_t$ with $\nabla \cdot (u_t - v_t)$; their Flow and Divergence Matching (FDM) augments the CFM objective with a tractable conditional surrogate estimated via Hutchinson's trace estimator.

DS-RectFlow shares the diagnosis that nonzero divergence signals regions where trajectories cross or compress, but differs in three ways. \textbf{First},we directly suppress $|\nabla\!\cdot\!\mathbf{v}|$ instead simply minimizing divergence. \textbf{Second}, the intervention is offline-only: it acts solely during coupling generation for RectFlow, leaving the training loss, architecture, and inference unchanged, so plain Euler inference runs at zero added latency, whereas FDS pays per-step overhead at every sampling call and FDM requires retraining with additional backward passes. \textbf{Third}, our framing is mechanistic: we treat divergence not only as a symptom but as a contributing cause of coupling crossings, and provide direct evidence (Appendix~\ref{sec:mechanism_chess}) that the \emph{convergent} component of $\nabla \cdot v$, rather than its unsigned magnitude, drives trajectory compression. These approaches are complementary, and combining correction at all three stages is left for future work.

\section{Divergence-Suppressing Flow Matching}
\label{sec:lpfm_section}
\begin{figure}[htbp]
  \centering
  \begin{subfigure}[t]{0.32\textwidth}
    \includegraphics[width=\linewidth]{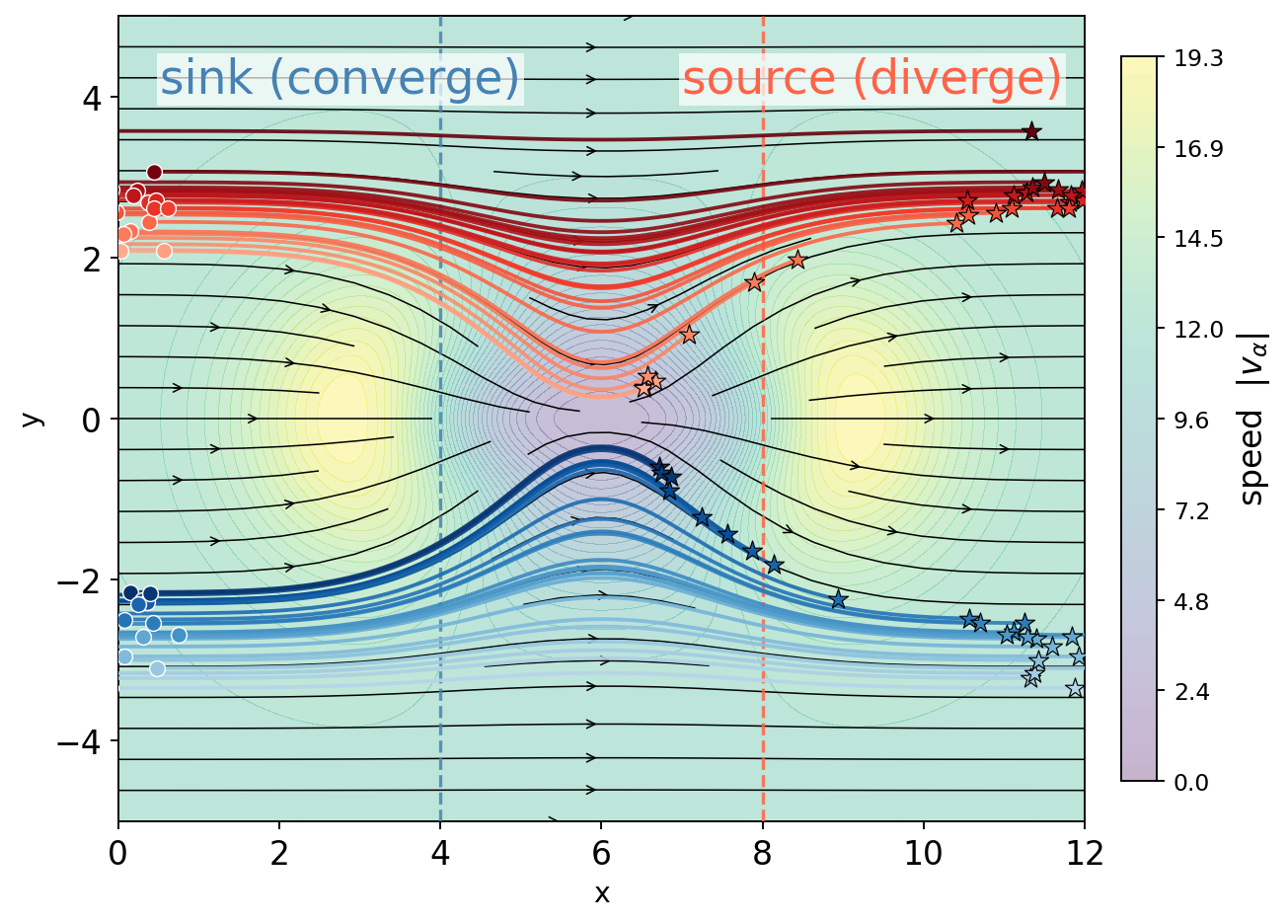}
    \caption{Loose budget: large $|\nabla\!\cdot\!v|$ near dipole.}
    \label{fig:leray_alpha_0}
  \end{subfigure}
  \hfill
  \begin{subfigure}[t]{0.32\textwidth}
    \includegraphics[width=\linewidth]{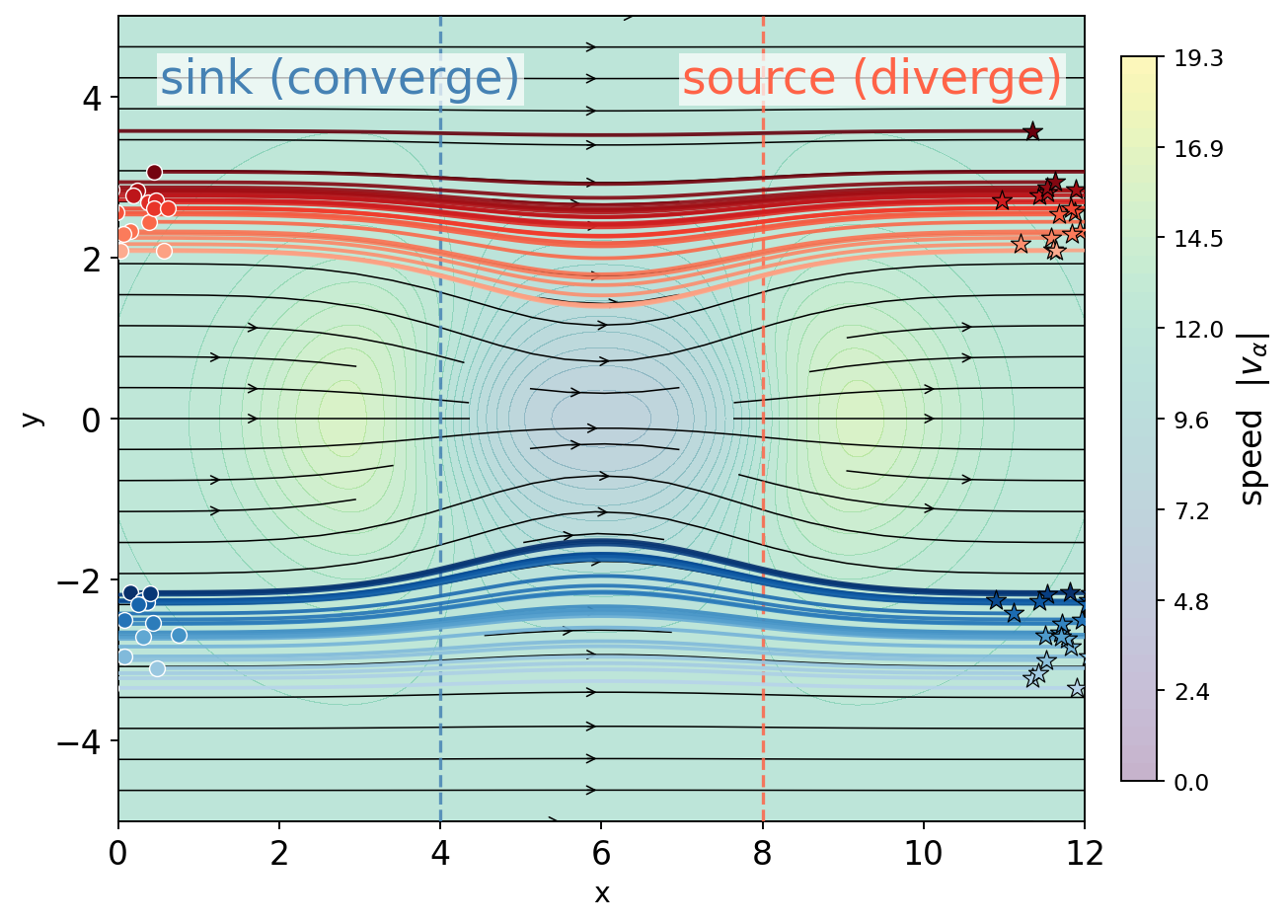}
    \caption{Intermediate budget: dipole partially suppressed.}
    \label{fig:leray_alpha_half}
  \end{subfigure}
  \hfill
  \begin{subfigure}[t]{0.32\textwidth}
    \includegraphics[width=\linewidth]{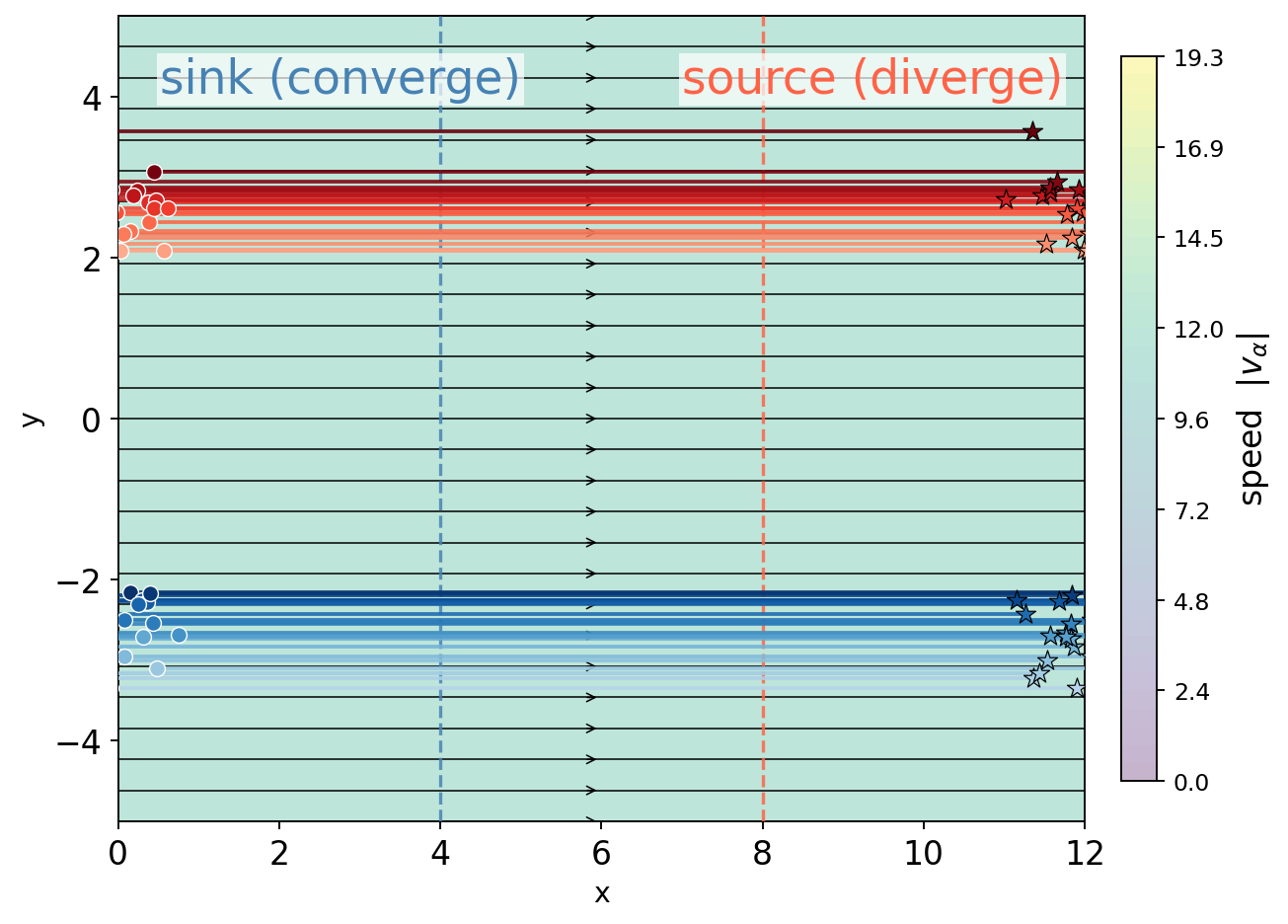}
    \caption{Tight budget: transport only.}
    \label{fig:leray_alpha_1}
  \end{subfigure}
  \caption{The same flow under three caps on $|\nabla\!\cdot\!v|$,
  with sources on the left and targets (stars) on the right. With
  a loose cap (a), a sink--source dipole bends the trajectories;
  tightening the cap (b, c) removes the dipole and the flow
  straightens into a clean rightward transport.}
  \label{fig:leray_alpha_sweep}
\end{figure}

Rectification needs a coupling whose trajectories do not bend. When
they do not, every intermediate state $x_t$ sees a single target
direction, the conditional expectation
$\E[x_1 - x_0 \mid x_t]$ in Eq.~\ref{eq:cfm_opt} picks it out
unambiguously, and the learned field is straight. When trajectories
arriving at the same $x_t$ disagree, the CFM target averages
conflicting directions and the field bends.
\begin{wrapfigure}{r}{0.42\linewidth}
  \vspace{-1em}
  \centering
  \begin{subfigure}[t]{\linewidth}
    \centering
    \includegraphics[width=\linewidth]{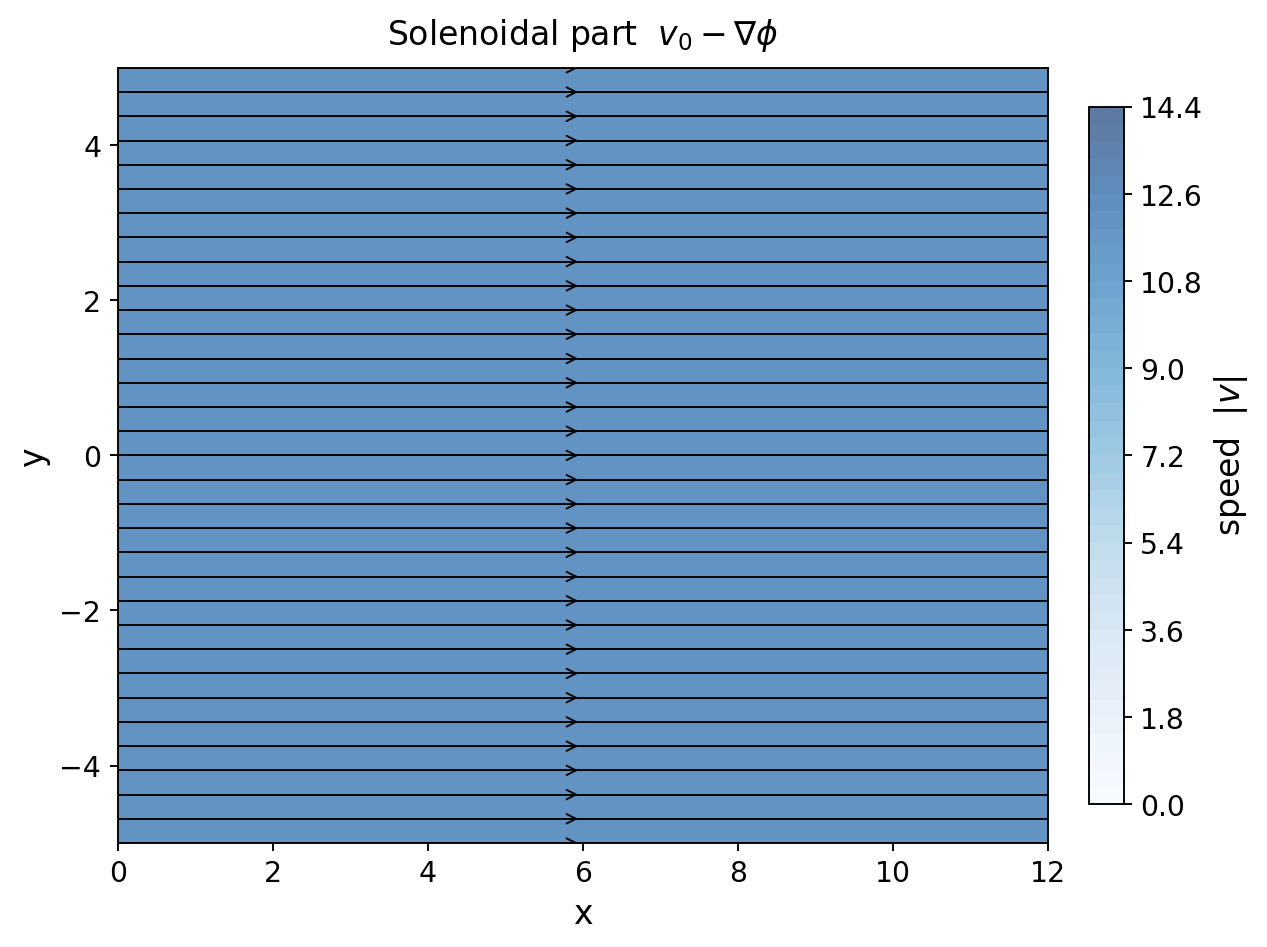}
    \caption{Transport part $u$, $\nabla\!\cdot u = 0$.}
    \label{fig:helmholtz_divfree}
  \end{subfigure}
  \\[0.5em]
  \begin{subfigure}[t]{\linewidth}
    \centering
    \includegraphics[width=\linewidth]{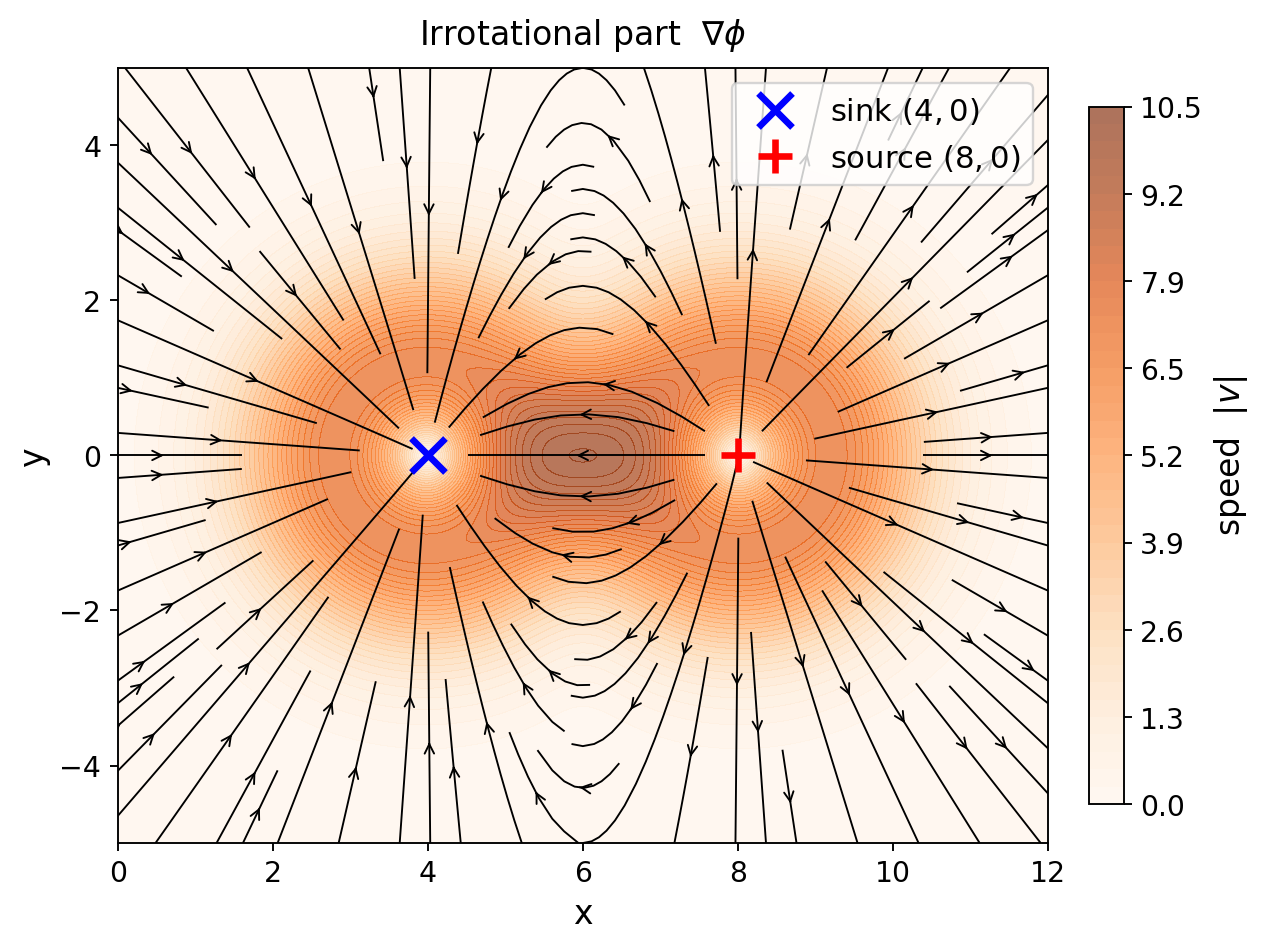}
    \caption{Dipole part $\nabla\phi$, carrying all of $\nabla\!\cdot\!v$.}
    \label{fig:helmholtz_div}
  \end{subfigure}
  \caption{Helmholtz decomposition of $v$ into a divergence-free
  transport (top) and a dipole (bottom) that carries all of its
  compressibility. Bending of trajectories happens only where the
  dipole is active.}
  \label{fig:helmholtz_decomposition}
  \vspace{-3em}
\end{wrapfigure}
\paragraph{Bending is compressibility.} Figure~\ref{fig:leray_alpha_sweep} shows a synthetic flow under
varying degrees of compressibility. The setup mirrors the toy
transport example used to motivate Rectified
Flow~\citep{liu2022rectifiedflow},\footnote{See the illustration at figure-2 in 
\url{https://rectifiedflow.github.io/blog/2024/intro/}.}
where two source bands must be transported to two target bands on
the opposite side of the domain.
Trajectories bend exactly where the velocity field compresses or
expands volumes. Where $\nabla\!\cdot\!v > 0$ neighbouring
trajectories spread apart; where $\nabla\!\cdot\!v < 0$ they get
pulled together and arrive at the same $x_t$. Either sign produces
the same effect on the coupling: nearby pairs $(x_0, x_1)$ that
should land at distinct endpoints are squeezed or fanned by the
field, and the resulting distortion propagates into the next round
of training. The local quantity that governs this is $|\nabla\!\cdot\!v|$,
and to keep trajectories straight it is enough to keep $|\nabla\!\cdot\!v|$ small.

\paragraph{Transport plus dipole.}
The Helmholtz decomposition makes this picture concrete. Any
sufficiently regular velocity field admits a unique split
\begin{equation}
  v \;=\; \underbrace{u}_{\text{transport}} \;+\; \underbrace{\nabla\phi}_{\text{divergence}},
  \qquad \nabla\!\cdot u \;=\; 0,
  \label{eq:helmholtz}
\end{equation}
where the transport part $u$ is divergence-free and accounts for
the bulk motion that moves probability mass between $\pi_0$ and
$\pi_1$, and the dipole part $\nabla\phi$ is irrotational and
carries all of the field's compressibility. Figure~\ref{fig:helmholtz_decomposition}
illustrates this split on the same  synthetic field as Figure~\ref{fig:leray_alpha_sweep}: the transport is a
smooth uniform motion, while the dipole concentrates near a single
sink--source pair and is small almost everywhere else. Bending of
trajectories is localized to the support of $\nabla\phi$;
everywhere $|\nabla\phi|$ is small, neighbouring trajectories
travel in parallel under $u$ alone.

\paragraph{Bounded compressibility.}
Demanding $\nabla\!\cdot\!v = 0$, i.e.\ killing the dipole
entirely, is too strong. By the continuity equation, any flow that
transports $\pi_0$ to a structured target $\pi_1$ with non-uniform
density must have nonzero divergence somewhere; an exactly
divergence-free integrator cannot reach $\pi_1$. What we actually
want is bounded compressibility along the integration path: a
budget $\epsilon(t) \ge 0$ such that
\begin{equation}
  |\nabla\!\cdot\!v(x,t)| \;\le\; \epsilon(t)
  \qquad\text{at every visited state } x.
  \label{eq:compressibility_budget}
\end{equation}
A coupling integrator that respects Eq.~\eqref{eq:compressibility_budget}
keeps the local Jacobian determinant of the flow map bounded by
$\exp(\int_0^t \epsilon(s)\,ds)$, so neighbouring trajectories
neither collapse nor explode by more than this factor. This is the
property that prevents coupling distortion: it is enough that
volumes change slowly, not that they are preserved exactly.
Tightening $\epsilon$ pushes the integrator further from the
high-$|\nabla\phi|$ regions of the field; loosening it recovers
vanilla Euler in the limit.

Figure~\ref{fig:helmholtz_decomposition} also tells us where the
budget actually matters. The dipole is small: it concentrates the
field's compressibility into a narrow region, while the rest of the
domain is dominated by the transport part $u$ and already satisfies Eq.~\eqref{eq:compressibility_budget} on its own. The budget only bites
in this narrow region, so enforcing it does not require touching the
field everywhere, only nudging trajectories away from the few states
where $|\nabla\!\cdot\!v(x,t)| > \epsilon(t)$. 
We apply this
correction at each Euler step during offline coupling generation
only; the next-round model is trained with the standard CFM loss
and run at inference with plain Euler, so the divergence-control
machinery adds no cost at sampling time.

\paragraph{Background: Avoiding High-Divergence Regions}
\label{sec:leray_bg}

\begin{wrapfigure}{r}{0.5\textwidth}
  \centering
  \includegraphics[width=0.5\textwidth]{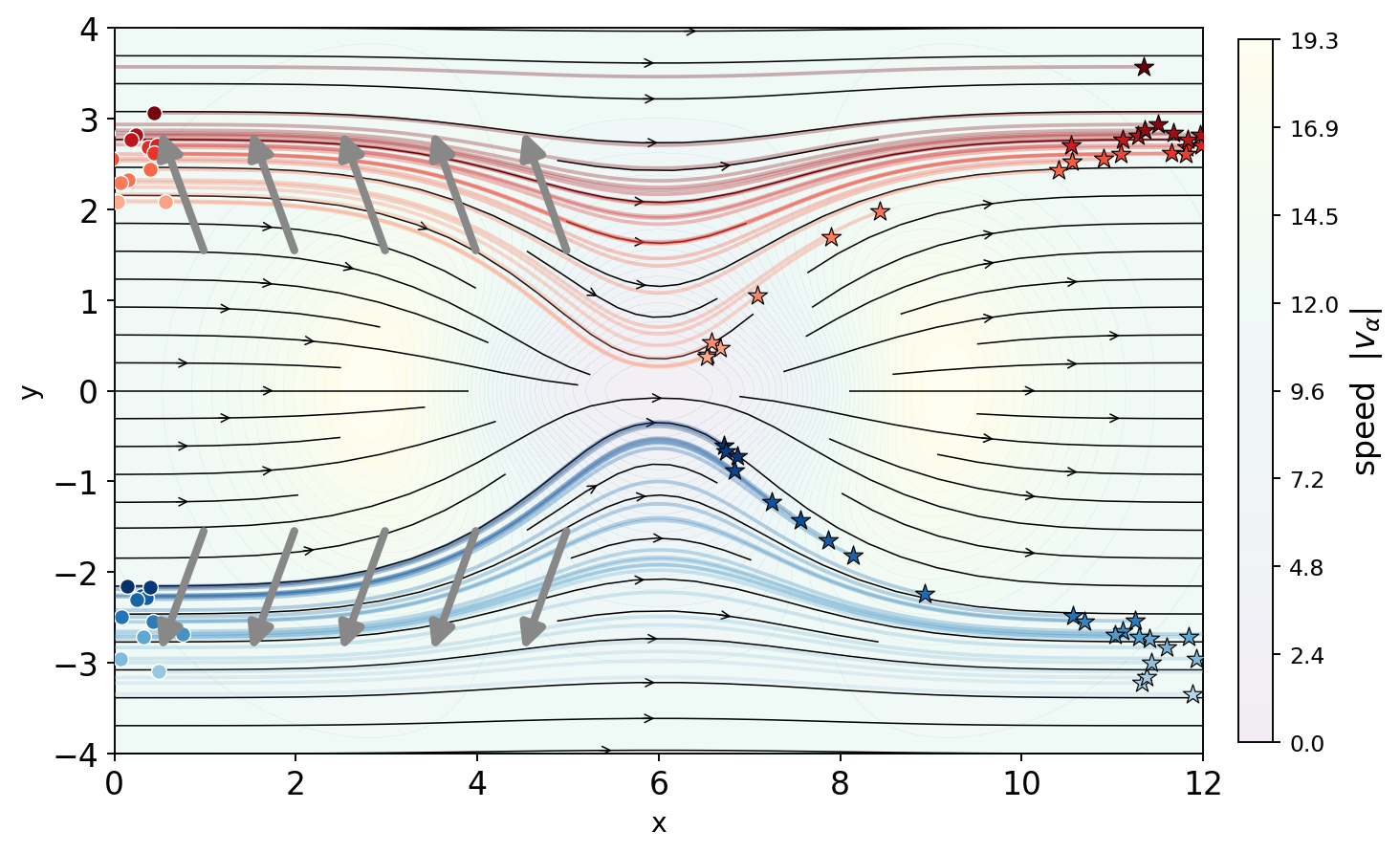}
  \caption{Pushing samples toward low-$|\nabla\!\cdot\! v|$ regions.}
  \label{fig:simple-push}
  \vspace{-2em}
\end{wrapfigure}

The mechanism we want is visible in Figure~\ref{fig:simple-push}:
trajectories that pass through high-$|\nabla\!\cdot\!\mathbf{v}|$
regions are pulled inward (where $\nabla\!\cdot\!\mathbf{v}<0$) or
pushed apart (where $\nabla\!\cdot\!\mathbf{v}>0$), bending the path
and producing distorted endpoints. If a particle can instead be
routed through a nearby in-budget corridor where
$|\nabla\!\cdot\!\mathbf{v}| \le \epsilon(t)$, it traces a cleaner
trajectory and yields a better $(x_0, x_1)$ coupling pair.

We control compressibility by acting on the state rather than the
field. At each Euler step, we displace the particle to a nearby
state $x^\star$ where $|\nabla\!\cdot\!\mathbf{v}(x^\star, t)|$ is
smaller, then integrate the unmodified velocity
$\mathbf{v}(x^\star, t)$. The velocity field itself is left
untouched; what changes is which part of the field the trajectory
samples. So long as $x^\star$ lies in the
$|\nabla\!\cdot\!\mathbf{v}| \le \epsilon(t)$ sub-level set, the
integrator advances under a velocity that is locally near-isochoric,
and the bound on the Jacobian determinant from
Eq.~\eqref{eq:compressibility_budget} carries over to the realized
trajectory. Operating on $|\nabla\!\cdot\!\mathbf{v}|$ rather than
on the signed divergence makes the correction sign-agnostic: it
suppresses both expansion and contraction, since either sign bends
trajectories.

\paragraph{Computing the Divergence in High Dimensions}
\label{sec:hutchinson}

To realize the surrogate we still need access to
$\nabla\!\cdot\!\mathbf{v}$, which is itself expensive to form
exactly in high dimensions. We therefore estimate it stochastically
via Hutchinson's trace estimator
\citep{hutchinson1989stochastic,grathwohl2018ffjord}, a detailed discussion of Hutchinson's trace estimator can be found in Appendix~\ref{app:hutchinson}.

We search for
a nearby state $x^\star$ at which
$|\widehat{\nabla\!\cdot}\,v_\theta(x^\star,t)|$ is reduced. At each
Euler step we draw $m$ Gaussian perturbations
$x'_j = x + \delta\,\xi_j$ ($\xi_j\sim\mathcal{N}(0,I)$) and select
the candidate minimising $|\widehat{\nabla\!\cdot}|$:
\begin{equation}
  x^\star = \operatorname*{arg\,min}_{x'\in\{x,\,x'_1,\ldots,x'_m\}}
            \bigl|\widehat{\nabla\!\cdot}\,v_\theta(x',t)\bigr|.
  \label{eq:accept}
\end{equation}
The corrected Euler step then advances from $x^\star$ rather than $x$:
\begin{equation}
  x_{t+\Delta t} =
  \begin{cases}
    x^\star_t + v_\theta(x^\star_t,\,t)\cdot\Delta t & t\leq t_{\mathrm{stop}},\\
    x_t + v_\theta(x_t,\,t)\cdot\Delta t & \text{otherwise.}
  \end{cases}
  \label{eq:hutchinson_euler}
\end{equation}
In other words, rather than projecting the velocity, we take the next
Euler step from a slightly displaced state whose velocity field is
closer to divergence-free in the Hutchinson estimate.
Algorithm~\ref{alg:hutchinson_search} details the per-sample batched selection.

\paragraph{Divergence-Suppressing RectFlow}
\label{sec:lpfm}
We now instantiate the divergence-suppressing correction inside the
rectification pipeline; we refer to the resulting method as
\textsc{DS-RectFlow}. The method
runs in parallel with vanilla RectFlow and shares its training
infrastructure; the only change is what is used to generate the
coupling. For each reflow round $k$, we draw $n$ sources
$x_0^{(i)} \sim \piZ$ and integrate each one forward for $N_\text{gen}$
steps using divergence-corrected Euler under the previous round's model
$v^{(k-1)}_\text{DS}$: at every integration step $(t, x)$, the
network's raw prediction $v_\theta(t, x)$ is replaced by its correction computed via
Algorithm~\ref{alg:hutchinson_search} before the particle takes a step. The
resulting endpoints $x_1^{(i)}$ form the coupling
$\mathcal{C}^{(k)}_\text{DS} = \{(x_0^{(i)}, x_1^{(i)})\}_{i=1}^n$.
We then train the next-round model $v^{(k)}_\text{DS}$ by
minimizing the standard CFM loss (Eq.~\ref{eq:cfm}) on this coupling, without changing the loss, architecture, or optimizer.  Intuitively, the correction nudges each particle toward a nearby
state with lower $|\nabla\!\cdot\! v|$.
Figure~\ref{fig:helmholtz_decomposition} makes this concrete: the
learned field decomposes into a divergence-free transport (panel a)
and a source/sink dipole (panel b) that carries all of its
divergence, and the corrected step pushes each particle away from
the dipole and onto the background transport.

At inference time, samples are generated by \emph{plain} Euler integration of the trained model and the divergence-suppressing correction does not appear anywhere in the sampling path. This is the central practical advantage of DS-RectFlow: the correction is a \emph{one-time offline cost}, paid once per coupling pair during data generation and amortized over the entire training run, while every subsequent inference call runs at \emph{identical wall-clock cost} to vanilla RectFlow. The method therefore inherits the inference profile of standard RectFlow, including its compatibility with single-step or few-step sampling, while delivering the cleaner couplings of a divergence-corrected integrator. In contrast to inference-time correction schemes such as higher-order solvers or guidance-based refinements, which trade per-step compute for sample quality, DS-RectFlow shifts the entire correction budget offline, so the improvements come for free at deployment.

\begin{algorithm}[ht]
\caption{\textsc{HutchinsonSearchCorrect}: low-$|\nabla\cdot v|$ neighbour search}
\label{alg:hutchinson_search}
\begin{algorithmic}[1]
\Require state $x$, time $t$, model $v_\theta$,
         scale $\delta$, candidates $m$, Hutchinson samples $n_h$
\Ensure  best state $x^\star$, velocity $v^\star$
\State $(v^\star,\hat{d}^\star)\leftarrow\textsc{HutchinsonDivEstimate}(x,t,n_h)$;\quad
       $x^\star\leftarrow x$
\For{$j=1,\ldots,m$}
  \State $x'\leftarrow x.\mathrm{detach}()+\delta\cdot\xi$,\quad $\xi\sim\mathcal{N}(0,I)$
  \State $(v',\hat{d}')\leftarrow\textsc{HutchinsonDivEstimate}(x',t,n_h)$
  \State $\mathrm{imp}_b\leftarrow|\hat{d}'_b|<|\hat{d}^\star_b|$ \quad for each $b$
         \hfill\Comment{push $|\nabla\cdot v|/d\to0$, both signs}
  \State update $x^\star,v^\star,\hat{d}^\star$ per sample where $\mathrm{imp}$
\EndFor
\State\Return $x^\star.\mathrm{detach}()$,\quad $v^\star.\mathrm{detach}()$
\end{algorithmic}
\end{algorithm}

\section{Experiments}
\label{sec:experiments}
We evaluate our method on 2D benchmarks and image generation, using
two related realizations of the divergence-suppressing correction.

\paragraph{2D benchmarks.} In low dimensions we can apply the
correction directly to the velocity at each Euler step. At state $x$
and time $t$, autograd gives us the velocity $v_0 = v_\theta(x,t)$
and its divergence $d_0 = \nabla \cdot v_\theta(x,t)$. We then
estimate the gradient of $|\nabla \cdot v_\theta|$ with one
finite-difference probe: sample a random unit direction $u$
uniformly on the unit sphere\footnote{$S^{d-1} =
\{u \in \mathbb{R}^d : \|u\|=1\}$ is the set of unit
vectors in $\mathbb{R}^d$.}, evaluate the divergence at the perturbed
point $x + h\, u$ for a small step $h$, and form
$\partial_u d \approx [(\nabla \cdot v_\theta)(x + h\, u) - d_0] / h$.
The rank-one estimate $\widehat{g} = (\partial_u d)\,u$ approximates
the gradient of the divergence in direction $u$, and the corrected
velocity is
\begin{equation}
    v_{\mathrm{proj}}
    \;=\;
    v_0 \;-\; \alpha \,
    \frac{\|v_0\|}{\|\widehat{g}\|}\,
    \mathrm{sign}(d_0)\,
    \widehat{g}.
\end{equation}

\paragraph{Image generation.} The first-order correction above is
infeasible at high dimension cases like images, so we use the zeroth-order surrogate of
Algorithm~\ref{alg:hutchinson_search}: at each step, $m$ candidate
states $x'_j = x + \delta \xi_j$ are drawn, and the Euler step
advances from the candidate with smallest Hutchinson-estimated
$|\nabla \cdot v|$. The strength parameters in the two settings are
not the same object: $\alpha \in [0,1]$ in the 2D case scales a
velocity correction, with $\alpha = 1$ matching the exact projection
of Figure~\ref{fig:leray_alpha_sweep}, whereas $\delta$ on images is
a positional search radius with no analogous limit. The two parameters control different mechanisms, $\alpha$ scales a
velocity correction in 2D, while $\delta$ is a positional search
radius on images, and never appear together in a single
experiment.
\subsection{Checkerboard Benchmark}
\label{sec:exp_chess}

We first apply the DS-RectFlow on Checkerboard benchmark, with set up detailed in Appendix~\ref{app:setup}.
We report the sliced Wasserstein distance (SWD) between a generated
sample and a held-out reference drawn from $\piOne$, using 2000
projections and $10{,}000$ samples per side, and the \emph{forbidden
cell fraction}, i.e.\ the fraction of generated points landing outside
the eight black cells. Both metrics are lower-is-better; SWD measures
overall distributional fidelity, while the forbidden cell fraction
isolates the mode of failure most specific to this benchmark---mass
leaking into regions that should have no density at all.
\begin{table}[ht]
  \centering
  \caption{%
    Checkerboard: SWD (lower is better) and inference time
    on $100$k samples, across NFE values. FM baselines fixed at
    NFE$=20$. Best SWD per NFE in bold. Rectflow methods share
    inference time at every NFE because the divergence-suppressing projection is
    offline-only.}
  \label{tab:chess_main}
  \setlength{\tabcolsep}{3pt}
  \small
  \begin{tabular}{l cc cc cc cc cc}
    \toprule
    & \multicolumn{2}{c}{NFE=1} & \multicolumn{2}{c}{NFE=5} & \multicolumn{2}{c}{NFE=10} & \multicolumn{2}{c}{NFE=15} & \multicolumn{2}{c}{NFE=20} \\
    \cmidrule(lr){2-3}\cmidrule(lr){4-5}\cmidrule(lr){6-7}\cmidrule(lr){8-9}\cmidrule(lr){10-11}
    Method & SWD & Time & SWD & Time & SWD & Time & SWD & Time & SWD & Time \\
    \midrule
    Flow Matching (ref)         & --  & --  & --  & --  & --  & --  & --  & --  & 0.162 & 0.066s \\
    \midrule
    RectFlow-$k{=}1$       & 0.1654 & 0.004s & 0.1662 & 0.017s & 0.1658 & 0.032s & 0.1653 & 0.048s & 0.1658 & 0.064s \\
    DS-RectFlow-$k{=}1$ & 0.1303 & 0.004s & 0.1263 & 0.017s & 0.1247 & 0.032s & 0.1236 & 0.048s & 0.1239 & 0.064s \\
    RectFlow-$k{=}2$       & 0.1657 & 0.004s & 0.1672 & 0.017s & 0.1664 & 0.032s & 0.1655 & 0.048s & 0.1669 & 0.064s \\
    \textbf{DS-RectFlow-$k{=}2$} & \textbf{0.1020} & 0.004s & \textbf{0.1014} & 0.017s & \textbf{0.1016} & 0.032s & \textbf{0.1006} & 0.048s & \textbf{0.1020} & 0.064s \\
    \bottomrule
  \end{tabular}
\end{table}
Table~\ref{tab:chess_main} reports SWD across five NFE values for
each method; Table~\ref{tab:chess_forbidden} reports the corresponding
forbidden cell fractions. Timings are measured on 100k samples with
the first batch excluded as warmup.
 \begin{figure}[htbp]
  \centering
  \begin{minipage}[t]{0.49\linewidth}
    \centering
    \includegraphics[width=\linewidth]{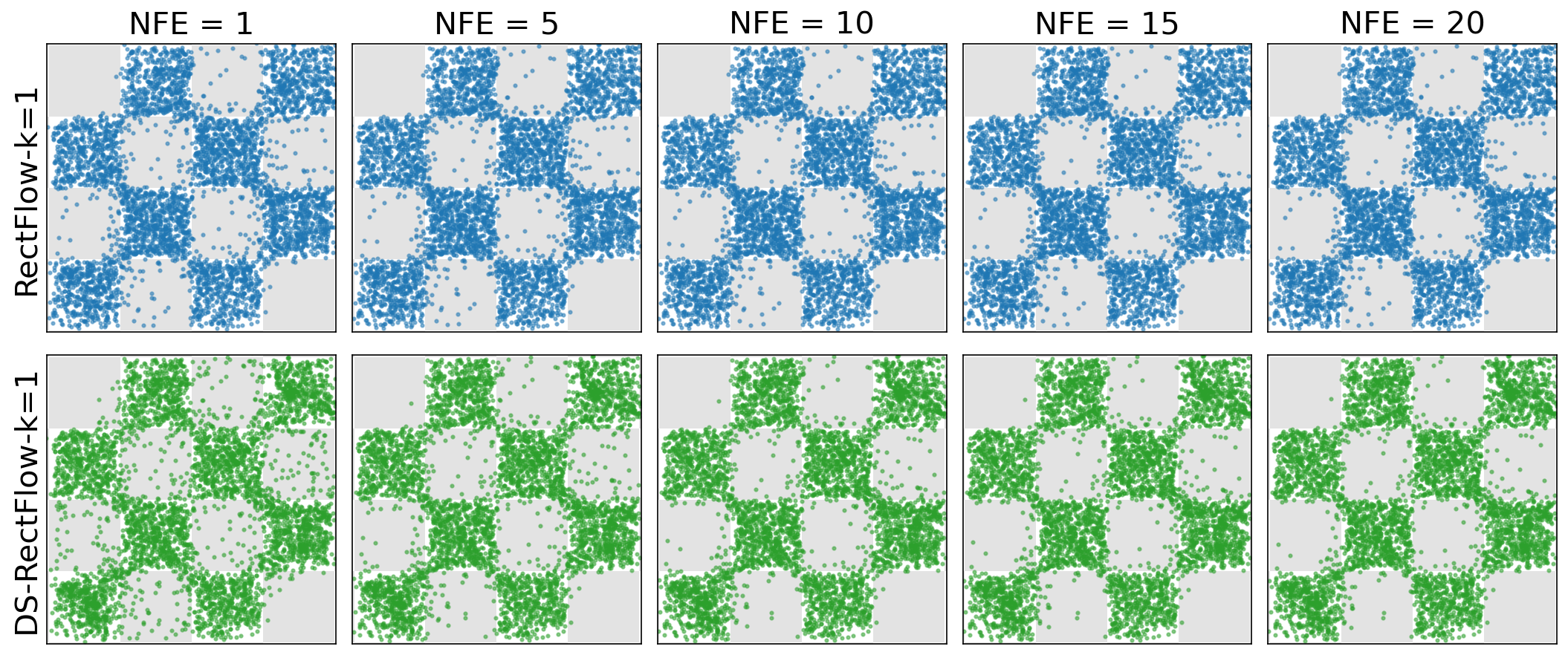}
    \caption{%
      Checkerboard, $k{=}1$. Top: RectFlow. Bottom:
      DS-RectFlow. Vanilla samples smear across cell boundaries;
      the divergence-suppressing variant has sharper cells but visible diagonal
      artefacts that one reflow round has not yet absorbed.
    }
    \label{fig:chess_qual_k1}
  \end{minipage}\hfill
  \begin{minipage}[t]{0.49\linewidth}
    \centering
    \includegraphics[width=\linewidth]{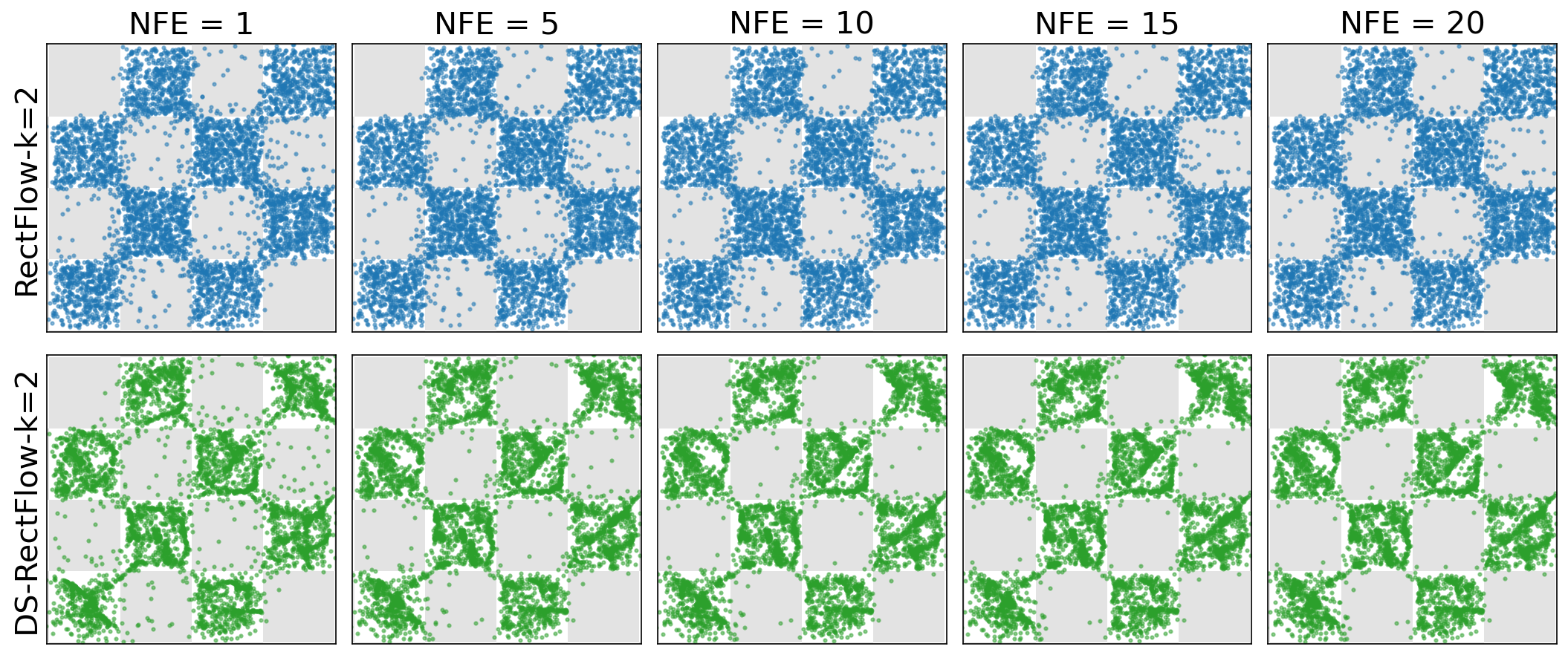}
    \caption{%
      Checkerboard, $k{=}2$. Top: RectFlow, essentially
      unchanged from $k{=}1$. Bottom: DS-RectFlow. The diagonal
      artefacts collapse into thin filaments, and panels are nearly
      identical across NFE, the signature of a straightened field.
    }
    \label{fig:chess_qual_k2}
  \end{minipage}
\end{figure}
Table~\ref{tab:chess_main} confirms the diagnosis from
Section~\ref{sec:rectflow}: the bottleneck is the integrator that
generates couplings, not the reflow objective. Vanilla RectFlow
stays near SWD $0.166$ at both $k{=}1$ and $k{=}2$, no better than
the FM baseline. The divergence-corrected integrator breaks this
plateau, reaching $0.124$ at $k{=}1$ and $0.102$ at $k{=}2$. A
cleaner coupling trains a cleaner model, which generates an even
cleaner coupling on the next round. This is the compounding
behaviour rectification is meant to produce, and it only appears
once the coupling is clean to begin with. The offline correction
also pays off at inference: DS-RectFlow-$k{=}2$ at NFE $=1$ runs
in $0.004$s and reaches SWD $0.102$, beating the $0.066$s
FM baseline by about $16\times$ in wall-clock. Its SWD is
flat across different NFE ($0.101$ to $0.102$), the signature of a
straightened field: 1 Euler step is as accurate as 20.

Table~\ref{tab:chess_forbidden} adds a complementary story. At
$k{=}1$, DS-RectFlow leaks \emph{more} mass into forbidden cells
than vanilla RectFlow at low NFE ($11.02\%$ vs.\ $7.85\%$ at NFE$=1$),
because its cleaner endpoints come with more complex intermediate
trajectories that one reflow round cannot reproduce in a single
step. The gap closes by $k{=}2$: DS-RectFlow at NFE$=1$ reaches
$4.99\%$, well
below vanilla RectFlow. At
NFE$=20$ it drops to $2.71\%$, less than half of the FM baseline.

\begin{wraptable}{r}{0.6\textwidth}
  \centering
  \vspace{-\baselineskip}
  \caption{%
    Checkerboard: Forbidden cell fraction (\%, lower is better).
    Fraction of generated samples landing outside the eight target
    black cells. FM baselines fixed at NFE$=20$.
  }
  \label{tab:chess_forbidden}
  \setlength{\tabcolsep}{1pt}
  \small
  \begin{tabular}{lccccc}
    \toprule
    Method & NFE=1 & NFE=5 & NFE=10 & NFE=15 & NFE=20 \\
    \midrule
    FM (ref, NFE=20)            & \multicolumn{5}{c}{6.06\% (fixed)} \\
    \midrule
    RectFlow-$k{=}1$            & 7.85\%  & 6.99\%  & 6.80\%  & 6.73\%  & 6.72\% \\
    DS-RectFlow-$k{=}1$      & 11.02\% & 7.08\%  & 5.99\%  & 5.53\%  & 5.33\% \\
    RectFlow-$k{=}2$            & 8.39\%  & 7.54\%  & 7.23\%  & 7.12\%  & 7.07\% \\
    \textbf{DS-RectFlow-$k{=}2$} & \textbf{4.99\%} & \textbf{3.46\%} & \textbf{3.06\%} & \textbf{2.87\%} & \textbf{2.71\%} \\
    \bottomrule
  \end{tabular}
\end{wraptable}

Figures~\ref{fig:chess_qual_k1} and~\ref{fig:chess_qual_k2} visualize the
results on the Checkerboard benchmark, with  each panel shows 100,000 generated samples overlaid on the checkerboard pattern. Vanilla RectFlow samples (blue) keep a
checkerboard envelope at every NFE and round, but mass bleeds into
the white cells along every boundary, most heavily at the
four-corner junctions.
DS-RectFlow (green) at $k{=}1$ has
sharper cells but visible diagonal bands of mis-routed mass through
the forbidden corners, the visual fingerprint of the trajectory
complexity above. By $k{=}2$ those bands collapse into thin
filaments, the eight black cells are cleanly separated with uniform
interior density, and the panels look nearly identical from NFE = 1
to NFE = 20, confirming that a single Euler step now suffices.

\subsection{GMM Crossing Benchmark}
\label{sec:exp_gmm}
We then apply the DS-RectFlow on Gaussian mixture model (GMM) benchmark, with set up detailed in Appendix~\ref{app:setup}. We report the SWD measures overall distributional fidelity to a
held-out reference sample from $\piOne$ together with inference time on $100$k samples.
\begin{table}[htbp]
\centering
\caption{GMM: SWD (lower is better) and inference time on $100$k samples, across NFE values. FM baselines fixed at NFE$=20$. Best SWD per NFE in bold. }
  \label{tab:gmm_main}
  \setlength{\tabcolsep}{3pt}
  \small
  \begin{tabular}{l cc cc cc cc cc}
    \toprule
    & \multicolumn{2}{c}{NFE=1} & \multicolumn{2}{c}{NFE=5} & \multicolumn{2}{c}{NFE=10} & \multicolumn{2}{c}{NFE=15} & \multicolumn{2}{c}{NFE=20} \\
    \cmidrule(lr){2-3}\cmidrule(lr){4-5}\cmidrule(lr){6-7}\cmidrule(lr){8-9}\cmidrule(lr){10-11}
    Method & SWD & Time & SWD & Time & SWD & Time & SWD & Time & SWD & Time \\
    \midrule
    FM (ref)          & --  & --  & --  & --  & --  & --  & --  & --  & 0.661 & 0.076s \\
    \midrule
    RectFlow-$k{=}1$       & 0.831 & 0.006s & 0.751 & 0.021s & 0.751 & 0.040s & 0.739 & 0.058s & 0.751 & 0.077s \\
    DS-RectFlow-$k{=}1$ & 1.755 & 0.006s & 0.888 & 0.021s & 0.606 & 0.040s & \textbf{0.533} & 0.058s & \textbf{0.501} & 0.076s \\
    RectFlow-$k{=}2$       & 0.801 & 0.006s & 0.790 & 0.021s & 0.785 & 0.039s & 0.782 & 0.058s & 0.788 & 0.077s \\
    DS-RectFlow-$k{=}2$ & \textbf{0.586} & 0.006s & \textbf{0.587} & 0.021s & \textbf{0.573} & 0.040s & 0.564 & 0.058s & 0.550 & 0.077s \\
    \bottomrule
  \end{tabular}
\end{table}
Table~\ref{tab:gmm_main} tells a variant of the checkerboard story
with one twist. On the checkerboard, DS-RectFlow helped at every
round and every NFE. On the GMM crossing benchmark, it is actually
\emph{worse} than vanilla RectFlow after a single round at low NFE
(SWD $1.755$ vs.\ $0.831$ at NFE$=1$, $k{=}1$). This is a diagnostic,
not a failure. The rotated-mode setting forces the velocity field to
decide which source mode is transported to which target, and the
DS-projected coupling captures this decision more faithfully than
vanilla Euler. The resulting $(x_0, x_1)$ pairs encode mode
assignment information that one round of reflow cannot yet reproduce
in a single step. The gap closes rapidly with more integration steps:
already at NFE$=10$ DS-RectFlow-$k{=}1$ ($0.606$) overtakes vanilla
RectFlow ($0.751$), and at NFE$=20$ it reaches $0.501$, well below both
the FM baseline ($0.661$) and vanilla RectFlow at any round.

By the second round the ordering flips even at NFE$=1$:
DS-RectFlow-$k{=}2$ ($0.586$) beats RectFlow-$k{=}2$ ($0.801$) by a
wide margin, and the gap holds across all NFE, widening to $0.550$ vs.\
$0.788$ at NFE$=20$.  The divergence-suppressing NFE$=1$ SWD drops from
$1.755$ at $k{=}1$ to $0.586$ at $k{=}2$  a
$3 \times$ improvement across rounds where vanilla RectFlow roughy keep the same (0.831 to 0.801).

DS-RectFlow-$k{=}2$ at
NFE$=20$ ($0.550$) and DS-RectFlow-$k{=}1$ at NFE$=20$ ($0.501$)
both achieve still lower SWD when more integration steps are
available, so the practical takeaway depends on the inference budget:
at NFE$=1$, $k{=}2$ is the best operating point; with more steps to
spare, fewer rounds suffice. Either way, DS-RectFlow models
match or beat the strongest inference-time alternative at a fraction
of the wall-clock cost.

\begin{figure}[htbp]
  \centering
  \begin{minipage}[t]{0.49\linewidth}
    \centering
    \includegraphics[width=\linewidth]{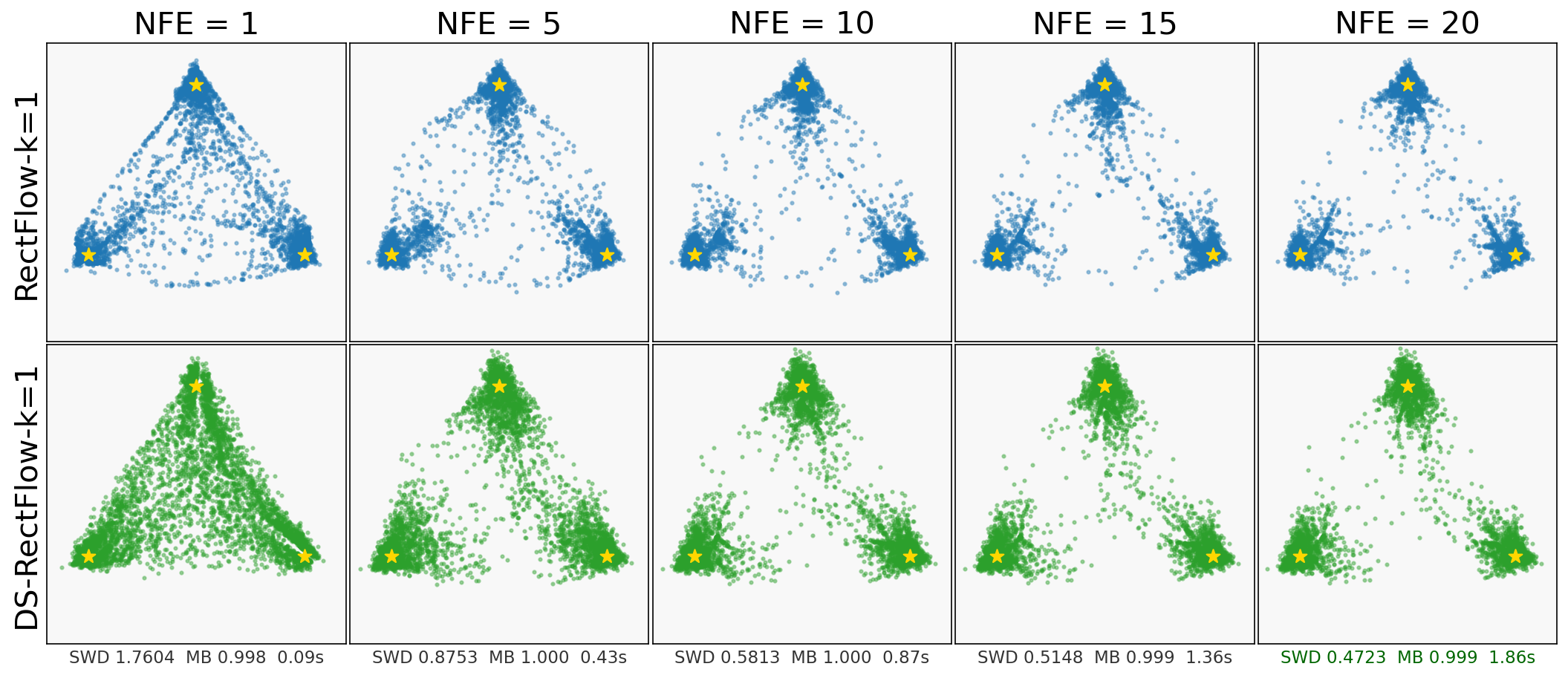}
  \caption{%
    GMM, $k{=}1$. Top: RectFlow. Bottom:
    DS-RectFlow. 
  }
    \label{fig:gmm_qual_k1}
  \end{minipage}\hfill
  \begin{minipage}[t]{0.49\linewidth}
    \centering
    \includegraphics[width=\linewidth]{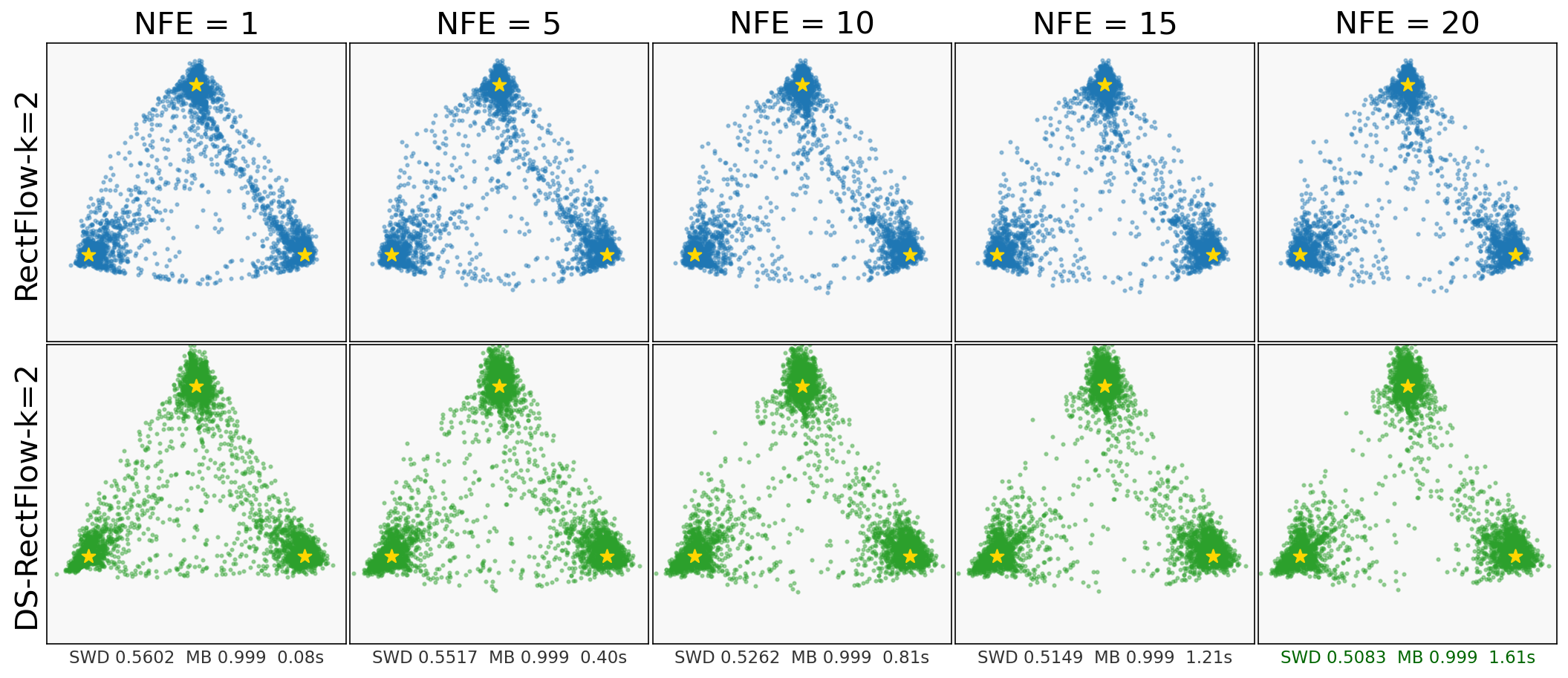}
  \caption{%
    GMM, $k{=}2$. Top: RectFlow. Bottom:
    DS-RectFlow. 
  }
    \label{fig:gmm_qual_k2}
  \end{minipage}
\end{figure}

Figure~\ref{fig:gmm_qual_k1} and \ref{fig:gmm_qual_k2} shows generated samples at $k{=} 1,2$ 
across NFE values $\{1,5,10,15,20\}$, with target modes marked by
gold stars and each panel showing 5,000 generated samples overlaid on
the GMM target. Both methods have largely cleaned up the central
crossing structure, but two differences remain visible. DS-RectFlow
clusters are tighter around each gold star and more evenly populated
across the modes as NFE increases, expecially when $k=2$, and the divergence-suppressing panels are noticeably more
invariant across NFE: the NFE$=1$ and $=20$ panels in the bottom
row look nearly indistinguishable, the same straightened-field
signature observed on the checkerboard.

\subsection{Cifar-10 Benchmark}
\label{sec:cifarexperiments}
We then apply the DS-RectFlow on Cifar-10, with setup detailed in
Appendix~\ref{app:setup}. We report Fr\'{e}chet Inception
Distance~(FID)~\citep{heusel2017gans} computed against the full
CIFAR-10 training set using $50{,}000$ generated samples and a fixed
torchvision InceptionV3 feature extractor, ensuring internal
consistency across all runs.
FID is evaluated at seven inference budgets,
$\mathrm{NFE}\in\{1,2,3,5,10,15,20\}$. We compare two offline pair
generation strategies: a fixed 20-step Euler solver and an adaptive
RK45 solver. In both cases, the vanilla and DS-RectFlow models use
plain Euler at inference; the divergence-suppressing correction is
applied \emph{only} during offline pair generation and incurs no
inference overhead.

In the Euler-20 setting we use $\delta=0.06$, and the check is
performed at each of the $N=20$ uniform steps that satisfy
$t\le t_{\text{stop}}$. In the RK45 setting we use $\delta=0.03$, and
the check is performed once per accepted adaptive step (rather than at
every internal RK45 stage); the  First Same As Last (FSAL) slope $f=v_\theta(x',t)$ is
recomputed after each accepted correction to keep the solver state
consistent. Beyond $t_{\text{stop}}$, both solvers revert to their
standard updates, Appendix~\ref{app:ablation-rk45-delta} and \ref{app:ablation-euler-delta} report the ablation study.

\begin{table}[h]
  \centering
  \caption{FID ($\downarrow$) on CIFAR-10 at
           varying inference NFE. Both models are 1-RectFlow
           ($k{=}1$) trained for $50{,}000$ iterations on top of a
           shared base FM model. We compare two offline pair
           generation solvers (Euler-20 and adaptive RK45). Inference uses plain Euler in
           all cases.}
  \label{tab:cifar10_fid_is}
  \setlength{\tabcolsep}{4pt}
  \begin{tabular}{llccccccc}
    \toprule
     & & \multicolumn{7}{c}{NFE} \\
    \cmidrule(lr){3-9}
    Offline solver & Method & 1 & 2 & 3 & 5 & 10 & 15 & 20 \\
    \midrule
    \multirow{2}{*}{Euler-20}
& RectFlow-$k{=}1$ (vanilla)
        & 17.41 & 16.26 & 15.59 & 15.21 & 14.60 & 14.71 & 14.38 \\
      & \textbf{DS-RectFlow-$k{=}1$ (ours)}
        & \textbf{12.03} & \textbf{10.45} & \textbf{9.93}
        & \textbf{9.53}  & \textbf{9.23}  & \textbf{9.10}
        & \textbf{8.97} \\
    \midrule
    \multirow{2}{*}{RK45}
      & RectFlow-$k{=}1$ (vanilla)
        & 9.13 & 8.10 & 7.56 & 7.24 & 6.93 & 6.79 & 6.65 \\
      & \textbf{DS-RectFlow-$k{=}1$ (ours)}
        & \textbf{8.94} & \textbf{7.46} & \textbf{7.01}
        & \textbf{6.64} & \textbf{6.37} & \textbf{6.40}
        & \textbf{6.17} \\
    \bottomrule
  \end{tabular}
\end{table}

Table~\ref{tab:cifar10_fid_is} reports FID at each NFE for
vanilla RectFlow-$k{=}1$ and DS-RectFlow-$k{=}1$ under both offline
solvers. DS-RectFlow achieves lower FID across every evaluated NFE
under both pair-generation regimes, and the higher-fidelity RK45
pairs yield substantially better FID overall for both methods.
Under the Euler-20 offline regime, DS-RectFlow shows large gains at
every NFE: at $\mathrm{NFE}\!=\!1$, FID drops from $17.41$ to
$12.03$ (a $31\%$ relative improvement); at $\mathrm{NFE}\!=\!10$,
from $14.60$ to $9.23$ ($37\%$); and at $\mathrm{NFE}\!=\!20$, from
$14.38$ to $8.97$ ($38\%$). Notably, DS-RectFlow at
$\mathrm{NFE}\!=\!1$ (FID $12.03$) already outperforms vanilla
RectFlow at every NFE in this sweep, demonstrating that the offline
coupling correction transfers directly into a single-step model
that is faster and better than 20-step inference on the vanilla
pipeline. Under the RK45 offline regime, both methods improve
markedly thanks to more accurate offline pairs, and DS-RectFlow
continues to lead at every NFE. At $\mathrm{NFE}\!=\!1$,
DS-RectFlow reduces FID from $9.13$ to $8.94$; at
$\mathrm{NFE}\!=\!5$, from $7.24$ to $6.64$; at
$\mathrm{NFE}\!=\!10$, from $6.93$ to $6.37$; and at
$\mathrm{NFE}\!=\!20$, from $6.65$ to $6.17$. Across both regimes,
the divergence-suppressing correction delivers consistent gains at
no inference cost, and the effect compounds with higher-quality
offline solvers.

\subsection{CelebA-64 Benchmark} \label{app:setup_celeba64} 
We further apply the DS-RectFlow on CelebA-64. We report FID  using $50{,}000$ generated samples. 
FID is evaluated at seven inference budgets,
$\mathrm{NFE}\in\{1,2,3,5,10,15,20\}$. We compare two offline pair
generation strategies: a fixed 50-step Euler solver and an adaptive
RK45 solver. In both cases, the vanilla and DS-RectFlow models use
plain Euler at inference; the divergence-suppressing correction is
applied \emph{only} during offline pair generation and incurs no
inference overhead. In the Euler-50 setting we use $\delta=0.03$, and the check is
performed at each of the $N=50$ uniform steps that satisfy
$t\le t_{\text{stop}}$. In the RK45 setting we use the setting  $\delta=0.03$. Training details can be found at Appendix~\ref{app:celeba64_training}.

\begin{table}[ht] 
\centering 
\caption{FID ($\downarrow$) on CelebA-64 at varying inference NFE. Both models are 1-RectFlow ($k{=}1$) trained for $1\!\times\!10^{5}$ iterations on top of a shared base FM model. We compare two offline pair generation solvers (Euler-50 and adaptive RK45) with $\delta{=}0.03$, $t_{\text{stop}}{=}0.5$. Inference uses plain Euler in all cases.} \label{tab:celeba64_fid} \setlength{\tabcolsep}{4pt} \begin{tabular}{llccccccc} \toprule & & \multicolumn{7}{c}{NFE} \\ \cmidrule(lr){3-9} Offline solver & Method & 1 & 2 & 3 & 5 & 10 & 15 & 20 \\ \midrule \multirow{2}{*}{Euler-50} & RectFlow-$k{=}1$ (vanilla) &  6.74 & 6.35 & 6.07 & 5.91 & 5.68 & 5.67 & 5.64 \\ & \textbf{DS-RectFlow-$k{=}1$ (ours)} & \textbf{5.87} & \textbf{4.90} & \textbf{4.79} & \textbf{4.46} & \textbf{4.23} & \textbf{4.08} & \textbf{4.08} \\ \midrule \multirow{2}{*}{RK45} & RectFlow-$k{=}1$ (vanilla) & 4.45 & 3.78 & 3.54 & 3.36 & 3.23 & 3.18 & 3.13 \\ & \textbf{DS-RectFlow-$k{=}1$ (ours)} & \textbf{4.27} & \textbf{3.58} & \textbf{3.49} & \textbf{3.28} & \textbf{3.09} & \textbf{3.04} & \textbf{3.02} \\ \bottomrule \end{tabular} 
\end{table}
Table~\ref{tab:celeba64_fid} reports FID at each NFE for vanilla RectFlow-$k{=}1$ and DS-RectFlow-$k{=}1$ on CelebA-64, under both Euler- and RK45-generated pair couplings. With Euler-generated pairs, DS-RectFlow achieves lower FID at every evaluated NFE, with gains that strengthen as NFE grows beyond one. At $\mathrm{NFE}\!=\!1$, FID drops from $6.74$ to $5.87$ (a $13\%$ relative improvement); at $\mathrm{NFE}\!=\!2$, from $6.35$ to $4.90$ (a $23\%$ improvement); at $\mathrm{NFE}\!=\!5$, from $5.91$ to $4.46$ ($25\%$); at $\mathrm{NFE}\!=\!10$, from $5.68$ to $4.23$ ($26\%$); and at $\mathrm{NFE}\!=\!20$, from $5.64$ to $4.08$ ($28\%$). Notably, DS-RectFlow at $\mathrm{NFE}\!=\!2$ (FID $4.90$) already outperforms vanilla RectFlow at every NFE in this sweep, so the offline coupling correction yields a two-step model that is both faster and sharper than $20$-step inference on the vanilla pipeline. With the more accurate RK45 generator, vanilla baselines tighten substantially (e.g.\ at $\mathrm{NFE}\!=\!1$, FID drops from $6.74$ under Euler-50 to $4.45$ under RK45), yet DS-RectFlow still wins at every NFE, with FID falling to $3.02$ at $\mathrm{NFE}\!=\!20$ versus $3.13$ for the vanilla counterpart. The trend mirrors what we observe on CIFAR-10: the divergence-suppressing correction delivers consistent gains at no inference cost, and the relative gap widens steadily as NFE increases before saturating near $\mathrm{NFE}\!\approx\!15$; we refer to the ablation studies in Appendix~\ref{app:ablation-euler-delta-celebA} and~\ref{app:ablation-rk45-delta-celebA}.
\section{Conclusion}
We presented DS-RectFlow, which replaces vanilla Euler with a
divergence-suppressing integrator during \textit{offline} coupling
generation. The correction removes the divergent component of the
learned velocity field and reduces coupling distortion, while
inference still runs plain Euler at the same wall-clock cost as
vanilla RectFlow. Across the checkerboard, GMM, CIFAR-10, CelebA-64
benchmarks, DS-RectFlow restores the compounding behaviour that
rectification promises but rarely delivers, and a single Euler step
on a divergence-suppressing model matches or beats many-step
alternatives. Our results identify coupling quality, not training
capacity or inference budget, as the principal bottleneck of
RectFlow, and show it admits a clean offline remedy. We expect the
same correction to help in stochastic interpolants, optimal-transport
FM, and score-based diffusion, wherever divergent drift components
degrade sample quality.

\bibliography{references}
\bibliographystyle{plainnat}
\newpage
\appendix
\section{Mechanism: Convergence Drives Trajectory Crossings}
\label{sec:mechanism_chess}

The gains reported above are consistent with the mechanism advanced in
Section~\ref{sec:lpfm}: trajectory crossings arise where the velocity
field locally compresses volume, and suppressing this compression at the
source yields cleaner couplings. We verify this mechanism directly on
the Checkerboard benchmark, and test whether it persists after
divergence suppression.

We probe each model at $t=0.5$, the time at which crossings concentrate
when the source and target have separated modes.
We draw $N=6400$ initial points $\{x_0^{(i)}\}$ i.i.d.\ from the
Gaussian source $\pi_0=\mathcal{N}(0,I)$ and evolve each under vanilla
Euler to obtain $x^{(i)}_{0.5}=\mathrm{Euler}(v,x_0^{(i)},20)$,
restricting the analysis to trajectories that the model actually visits
at inference time.
At each endpoint we compute two quantities.
The \emph{convergence field}
$c(x)=\max\bigl(0,-\nabla\!\cdot\!v(0.5,x)\bigr)$,
evaluated at $x=x^{(i)}_{0.5}$ via exact autograd, isolates the
instantaneously volume-contracting component of the velocity.
The \emph{trajectory compression score}
$s(x_0^{(i)})=\max\bigl(0,-\log|\det J_{x_0^{(i)}\to x^{(i)}_{0.5}}|\bigr)$,
where the flow-map Jacobian $J$ is estimated by finite differences with
$\varepsilon{=}0.05$ along the canonical axes, measures cumulative
volume contraction integrated along the trajectory.
The two are linked by the identity
$\tfrac{d}{dt}\log|\det J_{x_0\to x_t}| = \nabla\!\cdot\!v(t,x_t)$
along trajectories: if the negative divergence of $v$ is the dominant
driver of trajectory contraction, then $c$ and $s$ should be strongly
correlated when paired by their common spatial location $x^{(i)}_{0.5}$.

\subsection{Result}

\begin{figure}[htbp]
  \centering
  \includegraphics[width=\linewidth]{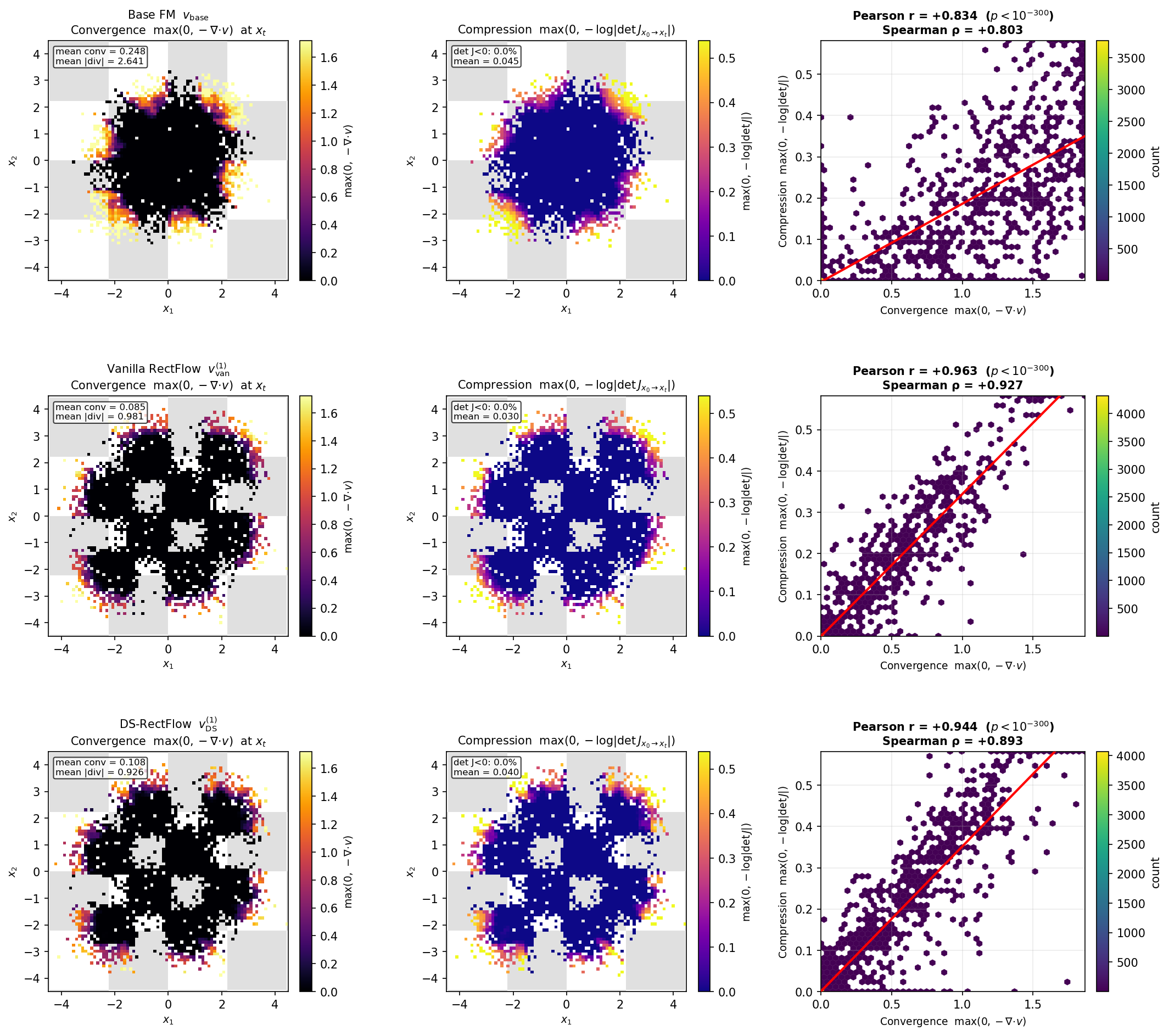}
  \caption{Divergence--compression analysis on $N=6400$ trajectories
    sampled from $\pi_0=\mathcal{N}(0,I)$ and evolved to $t=0.5$, for
    base FM (top), vanilla RectFlow (middle), and DS-RectFlow (bottom).
    \textit{Left:} convergence field $\max(0,-\nabla\!\cdot\!v)$ at
    $x_t$. \textit{Middle:} cumulative compression
    $\max(0,-\log|\det J_{x_0\to x_t}|)$.
    \textit{Right:} hexbin scatter with linear fit; Pearson
    $r=+0.834$, $+0.963$, $+0.944$ (all $p<10^{-300}$).
    Reflow concentrates convergence along the eight checkerboard cell
    boundaries and tightens its predictive link to compression
    ($r$ rises from $0.83$ to $\sim\!0.95$), while reducing
    $\langle|\nabla\!\cdot\!v|\rangle$ from $2.64$ to $\sim\!0.93$.
    DS-RectFlow lowers $\langle|\nabla\!\cdot\!v|\rangle$ further to
    $0.926$ without weakening the correlation, confirming the
    our divergence-suppressing correction acts in the intended direction
    (Table~\ref{tab:div_crossing}).}
  \label{fig:div_crossing_triptych}
\end{figure}

Figure~\ref{fig:div_crossing_triptych} reports the two spatial maps and
their joint distribution over the $6400$ paired observations for the
base FM model $v_\text{base}$.
The convergence field $c$ and compression score $s$ exhibit
Pearson $r=+0.8344$ ($p < 10^{-300}$) and Spearman $\rho=+0.8034$.
The two heatmaps share the same fine structure: bright regions of high
compression coincide with bright regions of high convergence.
We interpret this as direct empirical support for the causal chain that
motivates DS-RectFlow,
\begin{equation}
  \nabla\!\cdot\!v<0\;\Rightarrow\;
  \text{trajectory contraction}\;\Rightarrow\;
  \text{coupling crossings.}
\end{equation}

\subsection{Sign matters}

The mechanism is specific to the \emph{negative} component of
$\nabla\!\cdot\!v$.
Expansion ($\nabla\!\cdot\!v>0$) separates neighbouring trajectories and
reduces crossing risk; folding it into an unsigned magnitude cancels the
relevant signal.
The convergence field $c=\max(0,-\nabla\!\cdot\!v)$ isolates exactly
the part that drives crossings, which is why it yields a strong
correlation with compression while the unsigned divergence does not.

\subsection{Persistence across models}

Table~\ref{tab:div_crossing} reports the same statistics for vanilla
RectFlow $v^{(1)}_\text{van}$ and DS-RectFlow $v^{(1)}_\text{DS}$
after one reflow iteration.
Both reflow models cut the mean absolute divergence sharply
relative to the base ($\langle|\nabla\!\cdot\!v|\rangle$ drops from
$2.6413$ to $0.9813$ for vanilla and $0.9261$ for DS-RectFlow),
consistent with straighter flows that distort volume less.
DS-RectFlow gives a further ${\approx}6\%$ reduction beyond vanilla,
which is the direction the divergence-suppressing correction is
designed to push.
The convergence--compression correlation also strengthens after reflow
($r=+0.9632$ and $+0.9441$), since straighter trajectories concentrate
the remaining convergence into tighter regions that more cleanly
predict compression.
The high correlation for DS-RectFlow shows that the correction does
not break the underlying mechanism: there is less convergence
overall, but where it remains, it still drives compression.
Finally, none of the sampled trajectories cross at $t=0.5$, so
crossings are an off-distribution phenomenon in low-probability
regions of Gaussian space, exactly where the correction targets the
most severe contractions during coupling generation.

\begin{table}[htbp]
  \centering
  \small
  \begin{tabular}{lcccc}
    \toprule
    Model & Pearson $r$ & Spearman $\rho$
          & $\langle|\nabla\!\cdot\!v|\rangle$ & \% crossings \\
    \midrule
    Base FM $v_\text{base}$               & $+0.8344$ & $+0.8034$ & $2.6413$ & $0.0\%$ \\
    Vanilla RectFlow $v^{(1)}_\text{van}$ & $+0.9632$ & $+0.9270$ & $0.9813$ & $0.0\%$ \\
    DS-RectFlow $v^{(1)}_\text{DS}$       & $+0.9441$ & $+0.8926$ & $0.9261$ & $0.0\%$ \\
    \bottomrule
  \end{tabular}
  \caption{Divergence--compression correlation on $N=6400$ trajectories
    sampled from $\pi_0=\mathcal{N}(0,I)$, evaluated at $t=0.5$.
    Pearson $r$ and Spearman $\rho$ measure the correlation between
    convergence $\max(0,-\nabla\!\cdot\!v)$ and compression score
    $\max(0,-\log|\det J|)$.
    $\langle|\nabla\!\cdot\!v|\rangle$ is the mean absolute divergence,
    the quantity targeted by the divergence-suppressing correction.}
  \label{tab:div_crossing}
\end{table}

\section{Setup}
\label{app:setup}

\subsection{Checkerboard}
The target distribution $\piOne$ is a $4{\times}4$ checkerboard: a
uniform mixture over the eight black cells, each of area $(0.45)^2$,
with the remaining eight white cells forming \emph{forbidden regions}
where any generated mass represents an error. The source is a
standard Gaussian, $\piZ = \mathcal{N}(0, I_2)$. We use
$n = 200{,}000$ coupling pairs per reflow round,
$N_\text{gen} = 20$ Euler steps for coupling generation, and
$k = 2$ rectification rounds for both vanilla RectFlow and
DS-RectFlow; the divergence-suppressing correction strength is $\alpha = 0.5$. All
other training details are shared with the baselines.

\subsection{GMM}
\begin{figure}[ht]
    \centering
    \includegraphics[width=0.85\linewidth]{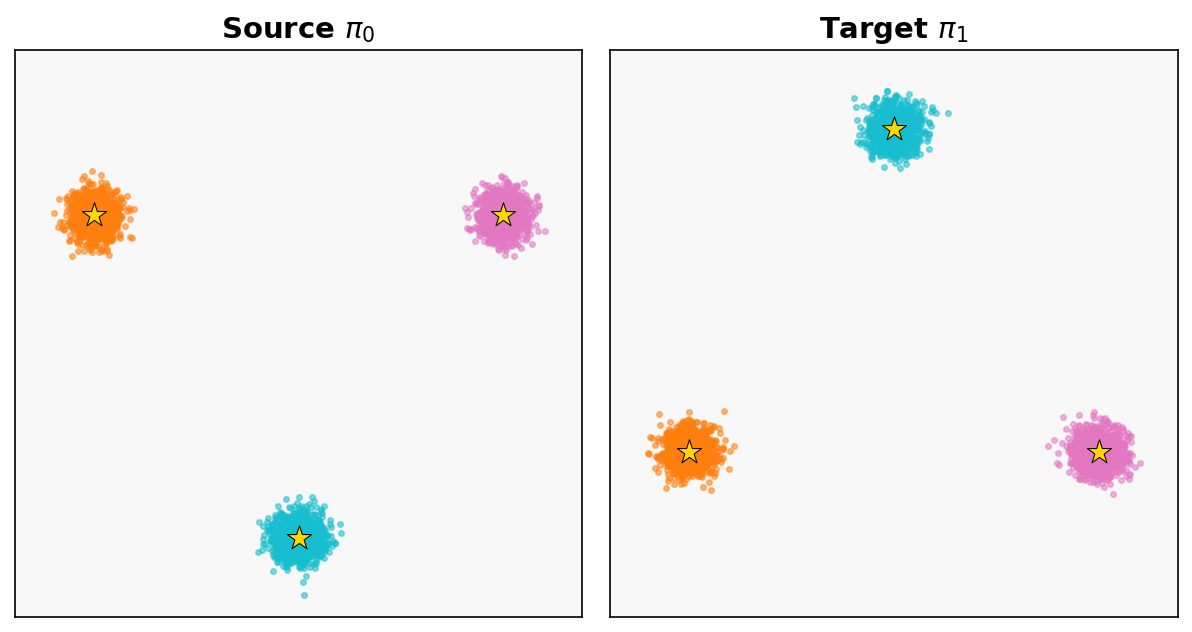}
    \caption{Source $\pi_0$ and target $\pi_1$ for the GMM crossing benchmark. Both are 3-mode Gaussian mixtures with $\sigma^2 = 0.3$ and modes on an equilateral triangle of circumradius $D = 10$; $\pi_1$ is $\pi_0$ rotated $60^\circ$ about the origin. Colors indicate the ideal mode-to-mode correspondence (orange$\to$orange, magenta$\to$magenta, cyan$\to$cyan), which forces all three transport paths through a congested central region and makes coupling crossings unavoidable under independent coupling.}
    \label{fig:gmm_ideal}
\end{figure}
Figure~\ref{fig:gmm_ideal} illustrates the GMM crossing benchmark. Both $\pi_0$
and $\pi_1$ are 3-mode Gaussian mixtures whose modes lie on an equilateral
triangle, with $\pi_1$ obtained by rotating $\pi_0$ by $60^\circ$ about the
origin. Under the colour-coded mode correspondence, each source cluster must
be transported to a target cluster on the opposite side of the centre, so the
three ideal transport paths are forced through a congested central region
where mode assignments compete. This makes the benchmark a canonical stress
test for RectFlow: coupling quality is determined by how cleanly the velocity
field resolves these crossings near the centre, and residual divergence in
this region is exactly what DS-RectFlow is designed to suppress.

Formally, $\pi_0$ has component variance $\sigma^2 = 0.3$ and modes at the
vertices of an equilateral triangle of circumradius $D = 10$ centred at the
origin,
\begin{equation*}
  \left\{\!
    \bigl(\tfrac{D\sqrt{3}}{2}, \tfrac{D}{2}\bigr),\;
    \bigl(-\tfrac{D\sqrt{3}}{2}, \tfrac{D}{2}\bigr),\;
    \bigl(0, -\tfrac{D\sqrt{3}}{2}\bigr)
  \!\right\},
\end{equation*}
and $\pi_1$ places its modes at
$\{(\tfrac{D\sqrt{3}}{2}, -\tfrac{D}{2}),\, (-\tfrac{D\sqrt{3}}{2}, -\tfrac{D}{2}),\, (0, \tfrac{D\sqrt{3}}{2})\}$.
We run $k = 3$ rectification rounds with $n = 100{,}000$ coupling pairs per
round, $N_\text{gen} = 20$ Euler steps for coupling generation, and
$\alpha = 0.5$ for the divergence-suppressing correction.

Both Checkerboard and GMM benchmarks use the same MLP architecture:
\begin{equation}
  v_\theta(t, x) = \text{MLP}([x;\, t]),
\end{equation}
where $[x;\, t] \in \R^3$ concatenates the 2D position $x$ with the scalar time $t$.
The MLP has three hidden layers of width 512 with ReLU activations.
Total parameters: approximately 800k.

\subsection{Cifar-10}
We evaluate on CIFAR-10~\cite{krizhevsky2009learning} ($32\!\times\!32$
RGB images, 50{,}000 training samples). Following the RectFlow
framework, we adopt a two-stage pipeline. In \emph{Stage~1} (base
model), we train a Flow Matching model with independent Gaussian
coupling using the NCSNpp U-Net architecture~\cite{song2020score}
($n_f\!=\!128$, channel multipliers $(1,2,2,2)$, 4 residual blocks
per resolution, attention at $16\!\times\!16$) for 50{,}000 iterations
with batch size 128, Adam optimiser ($\text{lr}\!=\!2\!\times\!10^{-4}$,
$\beta_1\!=\!0.9$), and EMA decay $0.9999$. In \emph{Stage~2} (reflow),
we generate 50{,}000 $(z_0,z_1)$ coupling pairs from the base model
and train a new model on those pairs for a further 50{,}000 iterations
under the same training configuration; this yields a 1-Rectified Flow
model in our $k$-counting convention ($k{=}1$).
For DS-RectFlow, the divergence-suppressing Euler integrator is
used during Stage~2 pair generation with hyperparameters
$\delta\!=\!0.05$, $t_{\mathrm{stop}}\!=\!0.5$, and 20 Euler steps.
For the vanilla RectFlow baseline, pairs are generated with
standard Euler integration under otherwise identical settings.
All experiments are run on a single NVIDIA H100 (80\,GB) GPU.

\paragraph{Comparison scope.}
Our absolute FID numbers are above those reported by \citet{liu2022rectifiedflow}; we train for  fewer iterations 50,000 instead of 400,000.
Our goal is \emph{not} to match the state-of-the-art absolute FID,
but to isolate the effect of the divergence-suppressing  coupling: both vanilla and
DS-RectFlow share \emph{identical} architectures, training
budgets, and hyperparameters, so every gap in
Table~\ref{tab:cifar10_fid_is} is attributable solely to the divergence-suppressing correction applied during offline pair generation.

\subsection{CelebA-64 Training Setup}
\label{app:celeba64_training}

This section gives the complete training and inference recipe used for
all CelebA-64 results in the paper. The pipeline has three stages:
(i) base 1-RectFlow pre-training, (ii) offline pair generation
(vanilla and divergence-suppressed), and (iii) reflow training on the
cached pairs. All stages share the same backbone, optimizer and data
preprocessing; only the iteration budget, dataset and (in stage~2)
the sampling solver change.

\paragraph{Dataset and preprocessing.}
We use the standard CelebA aligned/cropped split, resized to
$3{\times}64{\times}64$, with the official partition file giving
$162{,}770$ training images. Inputs are centered to $[-1,1]$ and
augmented with random horizontal flips during training. FID is
evaluated against the CelebA-64 training set with $50{,}000$ generated
samples using the torchvision InceptionV3 feature extractor.

All stages use the same NCSN++ / DDPM++ network, configured as a
$4$-scale variant for $64{\times}64$ inputs: channel multipliers
$(1,2,2,4)$ over resolutions $64{\to}32{\to}16{\to}8$, base width
$\text{nf}{=}128$, $2$ residual blocks per scale, BigGAN-style
residual blocks with skip rescaling, attention at the $16{\times}16$
feature map, group normalization with Swish activations, dropout
$0.1$, positional time embedding with Fourier scale $16$,
$3{\times}3$ convolutions and EMA decay $0.9999$. The reflow stage
reuses the stage-1 weights as initialization and keeps the
architecture identical.

We use Adam with learning rate $2{\times}10^{-4}$,
$\beta_{1}{=}0.9$, $\varepsilon{=}10^{-8}$ and zero weight decay. A
linear learning-rate warmup of $5{,}000$ iterations is applied at the
start of each stage, followed by a constant schedule. Gradients are
globally clipped at $1.0$. Mini-batch size is $128$ in every stage.
The rectified-flow loss uses uniform $t\!\in\![0,1]$ and an $\ell_{2}$
regression target with batch-mean reduction.

\paragraph{Stage 1: 1-RF pre-training.}
We train the base $1$-RectFlow on the CelebA-64 training set for
$5{\times}10^{5}$ iterations with the optimizer above. At batch size
$128$ over $162{,}770$ training images this is $\approx\!1{,}272$
iterations per epoch, i.e.\ $\approx\!393$ epochs in total. A snapshot
is written every $5{\times}10^{4}$ iterations (snapshots
$1,2,\ldots,10$) and a preemption-safe checkpoint every $10^{4}$
iterations. The stage-1 model used as the shared starting point for
both vanilla and DS-RectFlow branches is the final snapshot at
iteration $5{\times}10^{5}$.

\paragraph{Stage 2: offline pair generation.}
Starting from the stage-1 checkpoint, we generate $50{,}000$
$(\mathbf{z}_0,\mathbf{z}_1)$ pairs by integrating the learned
velocity field from $\mathbf{z}_0\!\sim\!\mathcal{N}(0,I)$ to
$\mathbf{z}_1$. Two solvers are reported:
\begin{itemize}
  \item \textbf{Euler-50}: a fixed $50$-step explicit Euler integrator.
  \item \textbf{RK45}: an adaptive Runge--Kutta--Fehlberg integrator
        (\texttt{scipy.integrate.solve\_ivp}) with absolute and
        relative tolerances both set to $10^{-3}$.
\end{itemize}
For DS-RectFlow, the divergence-suppressing correction is applied only
for $t\!\le\!t_{\text{stop}}$, with perturbation scale $\delta$,
$n_{H}{=}8$ Hutchinson probes per divergence estimate and $K{=}8$
candidate perturbations per step. We sweep $\delta$ and
$t_{\text{stop}}$ over the same grid as on CIFAR-10
($\delta\!\in\!\{0.01, 0.02, 0.03, 0.04, 0.05, 0.06\}$,
$t_{\text{stop}}\!\in\!\{0.2, 0.3, 0.4, 0.5\}$); the best operating
point on CelebA-64 is $\delta{=}0.03$, $t_{\text{stop}}{=}0.5$, used
for all reported numbers. The vanilla branch uses the same solvers
and the same $50{,}000$-sample budget but skips the DS correction.
Pair generation runs on a single H100 GPU with batch size $128$ and a
fixed random seed for reproducibility.

\paragraph{Stage 3: reflow training.}
On the cached pair dataset we train the reflowed model
($k{=}1$ reflow on top of the shared $1$-RF base) for $1{\times}10^{5}$
iterations using the same optimizer, batch size, EMA decay and
architecture as stage~1. The reflow loss is $\ell_{2}$ with a uniform
$t$-schedule. Snapshots are written every $10^{4}$ iterations and the
final ($10^{5}$-iter) checkpoint is used for all FID evaluations.
Vanilla RectFlow-$k{=}1$ baselines are trained with exactly the same
procedure on the vanilla pair dataset, so both rows of
Table~\ref{tab:celeba64_fid} differ only in the offline coupling.

\paragraph{Inference.}
At evaluation time we always use a plain Euler integrator, sweeping
$\mathrm{NFE}\!\in\!\{1, 2, 3, 5, 10, 15, 20\}$ on a uniform time grid
in $[0,1]$. Samples are generated from $\mathcal{N}(0,I)$ with initial
noise scale $1$ and decoded back to $[0,1]$ before FID computation.
Each FID number is computed from $50{,}000$ samples against the
CelebA-64 training reference statistics.

\paragraph{Compute.}
All stages are run on a single NVIDIA H100 (80\,GB) GPU. Stage~1
takes roughly four days; stage~2 takes hours (Euler-50) to about a
day (RK45 at $10^{-3}$ tolerance) per $(\delta, t_{\text{stop}})$
configuration; stage~3 takes about one day. End-to-end reproduction
of a single (Euler or RK45) row of Table~\ref{tab:celeba64_fid}
therefore requires $\sim\!6$ GPU-days from scratch, or
$\sim\!1.5$ GPU-days starting from the shared stage-1 checkpoint.

\section{Hutchinson-Based Divergence-Suppressing Correction: Full Details}
\label{app:hutchinson}

\subsection{Hutchinson Trace Estimator}

The divergence is the Jacobian trace:
\begin{equation}
  \operatorname{div} \mathbf{v}(x)
  = \mathrm{tr}\!\left(\tfrac{\partial \mathbf{v}}{\partial x}\right).
\end{equation}
Exact computation requires $d$ backward passes.  With Rademacher vectors
$\varepsilon \sim \{\pm 1\}^d$ (each coordinate independently $\pm 1$),
the Hutchinson estimator~\cite{hutchinson1989stochastic} gives an unbiased estimate:
\begin{equation}
  \mathrm{tr}(J) = \mathbb{E}_\varepsilon[\varepsilon^\top J\varepsilon]
  \approx \frac{1}{n_h}\sum_{s=1}^{n_h}\varepsilon_s^\top J\varepsilon_s.
\end{equation}
Each term $\varepsilon_s^\top J\varepsilon_s$ is computed in one VJP backward pass:
\begin{equation}
  \varepsilon_s^\top J\varepsilon_s
  = \Bigl(\nabla_x\bigl[v_\theta(x,t)\cdot\varepsilon_s\bigr]\Bigr)\cdot\varepsilon_s,
  \label{eq:vjp_detail}
\end{equation}
with \texttt{create\_graph=False}: the estimate is a scalar metric, never a
gradient direction, so no second-order graph is needed.
The normalised estimate used throughout is
\begin{equation}
  \widehat{\operatorname{div}} = \frac{1}{n_h d}
  \sum_{s=1}^{n_h}\varepsilon_s^\top J_{v_\theta}\varepsilon_s,
  \qquad \operatorname{Var}[\widehat{\operatorname{div}}] = O\!\left(\frac{1}{n_h d}\right).
\end{equation}

\subsection{Search-Based Correction Algorithm}

The corrected Euler step is then:
\begin{equation}
  x_{t_{i+1}} =
  \begin{cases}
    x^\star_{t_i} + v_\theta(x^\star_{t_i}, t_i)\cdot\Delta t & t_i \leq t_{\mathrm{stop}},\\
    x_{t_i} + v_\theta(x_{t_i}, t_i)\cdot\Delta t & \text{otherwise.}
  \end{cases}
\end{equation}

\subsection{Hyperparameters and Cost}

\begin{table}[ht]
\centering
\caption{Hutchinson hyperparameters (defaults used in all experiments).}
\label{tab:hparams_hutchinson}
\begin{tabular}{llll}
\toprule
Parameter & Symbol & Default & Role \\
\midrule
Perturbation scale & $\delta$ & 0.05 & Search radius around $x_t$ \\
Candidates & $m$ & 8 & Perturbed states evaluated per step \\
Hutchinson samples & $n_h$ & 8 & Variance reduction for $\widehat{\operatorname{div}}$ \\
Correction horizon & $t_{\mathrm{stop}}$ & 0.5 & Only correct for $t\leq t_{\mathrm{stop}}$ \\
Euler steps & $N$ & 20 & Steps for coupling generation \\
\bottomrule
\end{tabular}
\end{table}

\paragraph{Cost per corrected step.}
Each call to \textsc{HutchinsonSearchCorrect} requires $(m+1)$ calls to
\textsc{HutchinsonDivEstimate}, each costing $1 + n_h$ model passes (one forward,
$n_h$ VJP backward).  Total: $(m+1)(1+n_h)$ passes per corrected step.
For $m=n_h=8$: $81$ passes, applied to $\lceil t_{\mathrm{stop}}\cdot N\rceil = 10$
of 20 Euler steps.  No second-order computation graph is constructed at any point.

\section{Ablation Study: DS-RectFlow $\delta$ for RK45 Solver}
\label{app:ablation-rk45-delta}

We ablate the divergence scale $\delta$ used in the RK45 setting on CIFAR-10 reflow.
We evaluate five choices $\delta \in \{0.01, 0.02, 0.03, 0.04, 0.05\}$ and
report FID~$(\downarrow)$ after each reflow iteration $k \in \{1,2,3,5,10,15,20\}$.
The vanilla (no-projection) baseline is identical across all runs.

\begin{table}[ht]
\centering
\caption{%
  FID scores on CIFAR-10 for the vanilla RK45 baseline and DS-RectFlow  RK45
  across reflow iterations $k$, for different $\delta$.
  Bold indicates the best DS-RectFlow score at each $k$.
}
\label{tab:ablation-delta-rk45}
\setlength{\tabcolsep}{5pt}
\begin{tabular}{llccccccc}
\toprule
& & \multicolumn{7}{c}{Reflow iteration $k$} \\
\cmidrule(lr){3-9}
$\delta$ & Method & 1 & 2 & 3 & 5 & 10 & 15 & 20 \\
\midrule
    -
      & RectFlow-$k{=}1$ (vanilla)
        & 9.133 & 8.097 & 7.562 & 7.240 & 6.925 & 6.788 & 6.645 \\\midrule
{0.01}
  & DS-RectFlow-$k{=}1$    & 9.217 & 7.887 & 7.421 & 7.052 & 6.785 & 6.644 & 6.541 \\
\midrule
{0.02}
  & DS-RectFlow-$k{=}1$    & 9.059 & 7.711 & 7.279 & 6.963 & 6.615 & 6.395 & 6.383 \\
\midrule
0.03
  & DS-RectFlow-$k{=}1$    & \textbf{8.940} & \textbf{7.460} & \textbf{7.008} & \textbf{6.644} & \textbf{6.367} & 6.397 & \textbf{6.173} \\
\midrule
0.04
  & DS-RectFlow-$k{=}1$    & 8.839 & 7.397 & 6.995 & 6.666 & 6.450 & \textbf{6.251} & 6.275 \\
\midrule
0.05
  & DS-RectFlow-$k{=}1$    & 9.315 & 7.692 & 7.314 & 7.005 & 6.726 & 6.629 & 6.569 \\
\bottomrule
\end{tabular}
\end{table}

The results in Table~\ref{tab:ablation-delta-rk45} reveal a clear sweet spot
around $\delta = 0.03$.
A very small scale ($\delta=0.01$) triggers DS-RectFlow too conservatively:
the divergence rarely exceeds $0.01$ in well-trained regions, so few steps are
corrected and the gain over vanilla is modest (e.g.\ $6.541$ vs.\ $6.645$ at
$k{=}20$).
Increasing $\delta$ to $0.02$ and $0.03$ allows  to fire on a
broader set of steps, yielding larger FID improvements; $\delta=0.03$ achieves
the best final FID of $6.173$ at $k{=}20$.
At $\delta=0.04$, DS-RectFlow still provides gains but the best-iteration
score ($6.251$ at $k{=}15$) is slightly worse, and for $\delta=0.05$ the
benefit collapses almost entirely ($6.569$ vs.\ $6.645$ at $k{=}20$), suggesting
that a high threshold lets divergent steps pass uncorrected.
Based on this analysis we adopt $\delta=0.03$ for all RK45 experiments reported in the main paper.

\section{Ablation Study: DS-RectFlow $\delta$ for Euler Solver}
\label{app:ablation-euler-delta}
\begin{table}[ht]
\centering
\caption{%
  FID ($\downarrow$) on CIFAR-10 for the vanilla RectFlow-$k{=}1$ baseline and
  DS-RectFlow-$k{=}1$ across projection strengths~$\alpha$ at different NFE.
  Bold indicates the best DS-RectFlow score at each NFE.
}
\label{tab:ablation-alpha}
\setlength{\tabcolsep}{5pt}
\begin{tabular}{llccccccc}
\toprule
& & \multicolumn{7}{c}{NFE} \\
\cmidrule(lr){3-9}
$\delta$ & Method & 1 & 2 & 3 & 5 & 10 & 15 & 20 \\
\midrule
    -
      & RectFlow-$k{=}1$ (vanilla)
        & 17.408 & 16.264 & 15.585 & 15.207 & 14.595 & 14.712 & 14.378 \\\midrule
{0.01}
  & DS-RectFlow-$k{=}1$    & 17.220 & 16.141 & 15.248 & 14.845 & 14.390 & 14.175 & 14.227 \\
\midrule
{0.02}
  & DS-RectFlow-$k{=}1$    & 16.180 & 15.125 & 14.362 & 14.003 & 13.476 & 13.385 & 13.108 \\
\midrule
{0.03}
  & DS-RectFlow-$k{=}1$    & 14.957 & 13.822 & 13.174 & 12.623 & 12.278 & 12.035 & 12.072 \\
\midrule
{0.04}
  & DS-RectFlow-$k{=}1$    & 13.637 & 12.074 & 11.533 & 11.060 & 10.688 & 10.610 & 10.475 \\
\midrule
{0.05}
  & DS-RectFlow-$k{=}1$    & 12.313 & 10.683 & 10.143 &  9.735 &  9.308 &  9.244 &  9.208 \\
\midrule
0.06
  & DS-RectFlow-$k{=}1$
    & \textbf{12.029} & \textbf{10.448} & \textbf{9.933}
    & \textbf{9.534}  & \textbf{9.233}  & \textbf{9.100}
    & \textbf{8.972} \\
\midrule
{0.07}
  & DS-RectFlow-$k{=}1$    & 13.249 & 11.564 & 10.961 & 10.708 & 10.409 & 10.376 & 10.340 \\
\midrule
{0.08}
  & DS-RectFlow-$k{=}1$    & 16.531 & 14.832 & 14.280 & 14.224 & 14.031 & 13.841 & 13.717 \\
\midrule
{0.09}
  & DS-RectFlow-$k{=}1$    & 22.669 & 20.515 & 20.398 & 20.066 & 20.071 & 20.007 & 20.142 \\
\midrule
{0.10}
  & DS-RectFlow-$k{=}1$    & 31.885 & 29.149 & 29.046 & 28.672 & 29.054 & 29.309 & 29.213 \\
\bottomrule
\end{tabular}
\end{table}
\paragraph{Effect of projection strength~$\delta$.}
Table~\ref{tab:ablation-alpha} sweeps $\delta$ over one decade and reveals a clear unimodal structure.
At the two extremes the method fails for opposite reasons: very small $\delta$ (${\leq}0.02$) barely perturbs the coupling trajectories, so the learned velocity field is nearly identical to the vanilla baseline; very large $\delta$ (${\geq}0.09$) overwhelms the base velocity with the divergence correction, producing incoherent couplings and FID above the vanilla level (e.g.\ $31.9$ at $\delta{=}0.10$ versus $17.4$ for vanilla at NFE$=1$).

The optimal range is narrow but stable: $\delta \in [0.05, 0.07]$ all outperform vanilla at every NFE, with $\delta{=}0.06$ achieving the best score across all seven budgets.
At NFE$=1$, $\delta{=}0.06$ reduces FID from $17.41$ to $12.03$, a relative improvement of $31\%$.
The gain is largest at low NFE, where trajectory straightness matters most: the gap between $\delta{=}0.06$ and vanilla shrinks $5.4$ FID points at NFE$=1$ to NFE$=20$ ($14.38 \to 8.97$), remaining substantial throughout.
Notably, the performance within the stable band is flat enough that $\delta{=}0.05$  differs from the optimum by less than $0.3$ FID at NFE$=1$, indicating the method is not sensitive to the precise choice of $\delta$ as long as it stays in this range.

\section{Ablation Study: DS-RectFlow $\delta$ for Euler Solver}
\label{app:ablation-euler-delta-celebA}
\begin{table}[ht]
\centering
\caption{%
  FID ($\downarrow$) on CelebA-64 for the vanilla RectFlow-$k{=}1$ baseline and
  DS-RectFlow-$k{=}1$ across projection strengths~$\delta$ at different NFE,
  using the Euler solver with $t_{\text{stop}}{=}0.5$.
  Bold indicates the best DS-RectFlow score at each NFE.
}
\label{tab:ablation-alpha-euler-celebA}
\setlength{\tabcolsep}{5pt}
\begin{tabular}{llccccccc}
\toprule
& & \multicolumn{7}{c}{NFE} \\
\cmidrule(lr){3-9}
$\delta$ & Method & 1 & 2 & 3 & 5 & 10 & 15 & 20 \\
\midrule
    -
      & RectFlow-$k{=}1$ (vanilla)
        & 6.740 & 6.346 & 6.072 & 5.909 & 5.683 & 5.672 & 5.637 \\\midrule
{0.01}
  & DS-RectFlow-$k{=}1$    & 6.384 & 5.977 & 5.771 & 5.565 & 5.511 & 5.364 & 5.301 \\
\midrule
{0.02}
  & DS-RectFlow-$k{=}1$    & 6.149 & 5.483 & 5.265 & 4.970 & 4.724 & 4.731 & 4.725 \\
\midrule
{0.03}
  & DS-RectFlow-$k{=}1$
    & \textbf{5.872} & \textbf{4.900} & \textbf{4.785}
    & \textbf{4.456} & \textbf{4.227} & \textbf{4.084}
    & \textbf{4.076} \\
\midrule
{0.04}
  & DS-RectFlow-$k{=}1$    & 7.575 & 6.180 & 5.716 & 5.368 & 5.050 & 4.918 & 4.816 \\
\midrule
{0.05}
  & DS-RectFlow-$k{=}1$    & 13.760 & 11.078 & 10.391 & 10.001 & 9.641 & 9.421 & 9.306 \\
\bottomrule
\end{tabular}
\end{table}

\paragraph{Effect of projection strength~$\delta$ (Euler).}
Table~\ref{tab:ablation-alpha-euler-celebA} sweeps $\delta$ over the Euler-solver regime and reveals
a clearly unimodal structure.
At very small $\delta{=}0.01$ the projection barely perturbs the coupling trajectories, so
DS-RectFlow only modestly improves upon vanilla (e.g.\ $6.38$ vs.\ $6.74$ at NFE$=1$, a relative
improvement of just $5.3\%$); at very large $\delta{=}0.05$ the divergence correction overwhelms
the base velocity and the model collapses, with FID rising to $13.76$ at NFE$=1$, more than
double the vanilla baseline, and never recovering to vanilla levels even at NFE$=20$
($9.31$ vs.\ $5.64$).

The optimum sits cleanly at $\delta{=}0.03$, which dominates every NFE budget by a substantial
margin: at NFE$=1$ it cuts FID from $6.74$ to $5.87$ ($-12.9\%$), at NFE$=10$ from $5.68$ to
$4.23$ ($-25.6\%$), and at NFE$=20$ from $5.64$ to $4.08$ ($-27.7\%$).
The relative improvement \emph{grows} with NFE: Euler's discretisation error dominates the
straightening benefit at NFE$=1$, but as the step size shrinks the smoother DS-RectFlow
trajectories translate more directly into FID gains.
Surrounding values still beat vanilla; $\delta{=}0.02$ improves FID at every NFE (e.g.\
$4.73$ vs.\ $5.64$ at NFE$=20$), and $\delta{=}0.04$ beats vanilla once NFE${\geq}2$
(reaching $4.82$ at NFE$=20$).
However at NFE$=1$, $\delta{=}0.04$ already costs $1.7$ FID points relative to the optimum and
falls behind vanilla, indicating that the Euler solver is more sensitive to over-projection
than RK45 and that $\delta$ should be tuned conservatively when low-NFE generation is the
target use case.

\section{Ablation Study: DS-RectFlow $\delta$ for RK45 Solver}
\label{app:ablation-rk45-delta-celebA}
\begin{table}[ht]
\centering
\caption{%
  FID ($\downarrow$) on CelebA-64 for the vanilla RectFlow-$k{=}1$ baseline and
  DS-RectFlow-$k{=}1$ across projection strengths~$\delta$ at different NFE,
  using the adaptive RK45 solver with $t_{\text{stop}}{=}0.5$.
  Bold indicates the best DS-RectFlow score at each NFE.
}
\label{tab:ablation-alpha-rk45-celebA}
\setlength{\tabcolsep}{5pt}
\begin{tabular}{llccccccc}
\toprule
& & \multicolumn{7}{c}{NFE} \\
\cmidrule(lr){3-9}
$\delta$ & Method & 1 & 2 & 3 & 5 & 10 & 15 & 20 \\
\midrule
    -
      & RectFlow-$k{=}1$ (vanilla)
        & 4.453 & 3.778 & 3.542 & 3.358 & 3.230 & 3.177 & 3.127 \\\midrule
{0.01}
  & DS-RectFlow-$k{=}1$    & 4.455 & 3.768 & 3.505 & \textbf{3.221} & 3.163 & 3.048 & 3.158 \\
\midrule
{0.02}
  & DS-RectFlow-$k{=}1$    & 4.507 & 3.716 & 3.633 & 3.409 & 3.194 & 3.133 & 3.093 \\
\midrule
{0.03}
  & DS-RectFlow-$k{=}1$
    & \textbf{4.265} & \textbf{3.577} & 3.485
    & 3.275 & \textbf{3.088} & \textbf{3.036}
    & 3.021 \\
\midrule
{0.04}
  & DS-RectFlow-$k{=}1$    & 4.410 & 3.711 & 3.474 & 3.250 & 3.171 & 3.094 & 3.099 \\
\midrule
{0.05}
  & DS-RectFlow-$k{=}1$    & 4.450 & 3.776 & \textbf{3.465} & 3.315 & 3.141 & 3.175 & \textbf{2.930} \\
\bottomrule
\end{tabular}
\end{table}

\paragraph{Effect of projection strength~$\delta$ (RK45).}
Table~\ref{tab:ablation-alpha-rk45-celebA} sweeps $\delta$ over the adaptive RK45 regime.
In contrast to the Euler ablation, the FID surface is markedly flatter: all five DS-RectFlow
configurations stay within roughly $0.25$ FID of one another at every NFE budget, and even the
worst setting never falls more than $\sim$$0.07$ behind vanilla at any NFE.
This is consistent with the RK45 step-size controller absorbing much of the per-step error,
so the divergence-projection effect manifests as a smaller but more uniform shift.

Across NFEs the optimum sits around $\delta{=}0.03$, which achieves the best DS-RectFlow score
at four of the seven NFE budgets (NFE${\in}\{1,2,10,15\}$) and is essentially tied for the
remaining three.
At NFE$=1$ it cuts FID from $4.45$ to $4.27$ ($-4.1\%$), at NFE$=10$ from $3.23$ to $3.09$
($-4.4\%$), and at NFE$=15$ from $3.18$ to $3.04$ ($-4.4\%$).
The single largest gain in the table is $\delta{=}0.05$ at NFE$=20$, which reaches $2.93$
versus the vanilla $3.13$ ($-6.3\%$), but the same $\delta$ underperforms slightly at the
lowest NFE budgets, again pointing to the same low-NFE sensitivity to over-projection that
the Euler study revealed, only attenuated here by the adaptive stepping.

Compared to the Euler results in Table~\ref{tab:ablation-alpha-euler-celebA}, two patterns are worth
noting. First, the absolute gains are smaller under RK45 (around $0.1$ to $0.2$ FID) because
the vanilla baseline is already much stronger ($\sim$$3$ FID versus $\sim$$5$\textendash$7$ FID
for Euler), leaving less headroom for the projection to recover. Second, the optimal $\delta$
shifts upward from $0.02$ (Euler) to $0.03$ (RK45), suggesting that smoother solvers can
tolerate, and benefit from, a slightly stronger divergence correction.


\end{document}